%% file: sample_paper.tex
\documentclass[twoside]{article}

\usepackage[preprint]{aistats2026}
\usepackage[round]{natbib}

\input{math_commands.tex}

\usepackage{hyperref}
\usepackage{url}
\usepackage{xcolor}         
\usepackage{tikz,lipsum,lmodern}
\usepackage[most]{tcolorbox}
\usepackage{bm}
\usepackage{amssymb,amsmath,amsthm}
\usepackage{graphicx,ctable,booktabs}
\usepackage{caption}
\usepackage{subcaption}
\usepackage{enumerate}
\usepackage{comment}
\usepackage{algorithm,algpseudocode}
\usepackage{wrapfig}
\usepackage[font=scriptsize,labelfont=bf]{caption}

\newtheorem{observation}{Observation}
\newtheorem{proposition}{Proposition}
\newtheorem{lemma}{Lemma}

\newtheorem{assumption}{Assumption}
\newtheorem{theorem}{Theorem}

\newcommand{\kr}[1]{{\color{blue} [KR: {#1}]}}

\makeatletter
\newtheorem*{rep@theorem}{\rep@title}
\newcommand{\newreptheorem}[2]{%
\newenvironment{rep#1}[1]{%
 \def\rep@title{#2 \ref{##1}}%
 \begin{rep@theorem}}%
 {\end{rep@theorem}}}
\renewcommand{\eqref}[1]{\textup{(\ref{#1})}}
\makeatother

\newreptheorem{theorem}{Theorem}
\newreptheorem{lemma}{Lemma}
\newreptheorem{observation}{Observation}
\newreptheorem{proposition}{Proposition}

\begin{document}

%

%

\runningauthor{Kyung Rok Kim, Yumo Bai, Chonghuan Wang, Guanting Chen}

\renewcommand{\thefootnote}{\fnsymbol{footnote}}
\twocolumn[
\aistatstitle{Understanding the Impact of Sampling Quality in Direct Preference Optimization}

\aistatsauthor{ 
Kyung Rok Kim$^{1}$\footnotemark$^\dagger$ \And
Yumo Bai$^{1}$\addtocounter{footnote}{-1}\footnotemark \And 
Chonghuan Wang$^2$ \And 
Guanting Chen$^{1}$$^\dagger$
}

\aistatsaddress{ 
$^1$University of North Carolina at Chapel Hill \quad\quad 
$^2$University of Texas at Dallas \\ 
$^\dagger$\texttt{\{kkrok,guanting\}\@unc.edu} 
}
]

\footnotetext{Equal contribution}
\renewcommand{\thefootnote}{\arabic{footnote}}

\begin{abstract}
We study how data of higher quality can be leveraged to improve performance in Direct Preference Optimization (DPO), aiming to understand its impact on DPO’s training dynamics. Our analyses show that both the solution space and the convergence behavior of DPO depend on the support and quality of the data-generating distribution. We first analyze how data and reference policy influence policy updates during gradient descent, and how a practical phenomenon known as likelihood displacement can interfere with the desired dynamics. We then design a simplified yet well-structured alignment model as a proxy that preserves most of the beneficial properties of RLHF while avoiding likelihood displacement. Based on this model, we develop quantitative results showing how more frequent high-quality responses amplify the gradient signal and improve the optimization landscape, leading to more effective policy learning. Our theoretical findings are supported by empirical experiments and provide a principled justification for the online DPO framework in practice. 
\end{abstract}

\section{Introduction}

Reinforcement Learning from Human Feedback (RLHF) \citep{christiano2017deep, ouyang2022training} has become a cornerstone for aligning large language models (LLMs) with human preferences, playing a central role in the development of modern LLMs \citep{achiam2023gpt, guo2025deepseek}. RLHF trains models using human preference data with a learned reward model. While RLHF has demonstrated strong empirical performance and inspired numerous follow-up works \citep{lee2023rlaif, wu2023fine, azar2024general}, its training procedure is often expensive and lacks interpretability.
Almost in parallel, DPO \citep{rafailov2023direct} has emerged as a simpler and more interpretable alternative.
Instead of training a separate reward model, DPO directly optimizes a model to prefer one response over another. DPO obviates the need for reward models, which were often pointed out as the cause of challenges in RLHF \citep{gao2023scaling,miao2024inform}, and has inspired a considerable amount of interest \citep{xu2023some,ivison2024unpacking, tang2024generalized,qi2024online,dong2024rlhf}. 

Among the existing literature, there is a broad consensus that data for preference learning remains severely limited, with only a handful of available datasets. This has motivated both theoretical and empirical efforts to better utilize data in DPO and, more broadly, in RLHF. However, a clear gap persists between theory and practice. On the theoretical side, works such as \citet{shi2024crucial,feng2025pilaf,xiong2024iterative,tajwar2024preference,guo2024direct} study how sampling schemes in online data collection can accelerate convergence or guarantee desirable first-order conditions, while \citet{pan2025matters, won2025differential} investigate the role of data distributions from an optimization perspective. Although these analyses are statistically rigorous, they often rely on assumptions that are difficult to satisfy in practice. In contrast, the empirical literature proposes practical strategies for improving DPO performance, see, e.g., \citet{lee2023rlaif, pang2024iterative, pattnaik2024enhancing, kim2024sdpo, wu2024beta, chen2024towards, xiao2025finding}, but with limited theoretical justification. Efforts to bridge this gap have encountered unexpected new phenomena, most notably \textit{likelihood displacement} \citep{shi2024understanding, razin2024unintentional}, where the likelihood of both preferred and dispreferred responses decreases after DPO training. Further literature review is provided in Appendix \ref{app:related_work}.

\textbf{Contribution}. Motivated by these challenges, we adopt a modeling-based and practice-driven approach to analyze DPO dynamics. We aim to bridge theory and application by developing a model that disentangles the complexities of practical DPO training and makes its dynamics more interpretable.


\begin{itemize}
    \item We establish theoretical results that characterize how the support and quality of the generating distribution shape the optimization landscape. In particular, we rigorously examine how the data distribution governs per-step gradient dynamics and provide conditions under which the policy probability of a response increases or decreases. 
    
    \item We link per-step gradient analysis to the contradictory phenomenon of likelihood displacement. We present empirical evidence that highlights its connection to semantic correlations for preference pairs in the batch.
    
    \item To disentangle the complexities of practical DPO training, we design a simplified yet well-structured model for RLHF alignment that serves as an effective proxy for analyzing DPO training behavior, abstracting the essence from the complexities of full-scale LLM alignment. This model retains the convergence guarantees valued in theoretical analysis while being free from likelihood displacement. It thus offers a rigorous and clean platform for studying newly proposed algorithms and strategies without interference from semantic correlations or other unexpected behaviors.
    
    \item Based on this framework, we integrate our theoretical insights into the online DPO/RLHF framework and demonstrate the advantages of iterative training with improved sampling, particularly when using AI-generated feedback. Empirical results support our theoretical findings and validate the proposed framework.
\end{itemize}


\subsection{Preliminaries} \label{section:prelim}

Let $\mathcal{X}$ and $\mathcal{Y}$ denote the spaces of prompts and responses, respectively. Since our work does not involve learning the reward model, we assume access to an oracle reward function $r^*(\cdot,\cdot): \mathcal{X} \times \mathcal{Y} \to \mathbb{R}$ throughout the paper. Given a prompt $x \in \mathcal{X}$ sampled from a distribution $\mathcal{D}_X$, a policy $\pi_\theta(\cdot \mid x)$ defines a probability mass function (or a density, depending on the model) over $\mathcal{Y}$, parametrized by $\theta \in \Theta$. The goal of RLHF is to maximize the reward keeping $\pi_\theta$ close to a reference model $\pi_{ref}$, which is formulated as the RLHF objective:
\begin{equation}
\begin{aligned}\label{RLHF_original}
\max_{\theta\in\Theta}\mathbb{E}_{x\sim\mathcal{D}_X, y\sim\pi_{\theta}(y|x)}[r^*(x,y)]- \\
\beta \mathbb{D}_{KL}[\pi_{\theta}(y|x) || \pi_{\text{ref}}(y|x)].
\end{aligned}   
\end{equation}

As the RLHF objective is difficult to maximize in practice, \cite{rafailov2023direct} assumed the Bradley-Terry (BT) model and proposed to optimize DPO loss instead. For a prompt $x$ and a pair of responses $y_1,y_2$, the BT model ranks $y_w = y_1$ and $y_l = y_2$ if $y_1 \succ y_2$ (indicating that $y_1$ is more preferred than $y_2$), with the probability 
\begin{align*}
    \mathbb{P}(y_1 \succ y_2|x) &= \frac{\exp(r^*(x, y_1))}{\exp(r^*(x, y_1)) + \exp(r^*(x, y_2))} \\
    &= \sigma(r^*(x,y_1) - r^*(x,y_2)).
\end{align*}
Under this setting, the goal of DPO is to optimize the DPO loss
\begin{align}\label{DPO_BT_definition}
    &\mathcal{L}(\theta, \mathcal{D}) =-\mathbb{E}_{(x,y_w,y_l)\sim \mathcal{D}}\notag\\ 
    &\quad\;\left[\log \sigma\left(\beta \log \frac{\pi_{\theta}(y_w|x)}{\pi_{\text{ref}}(y_w | x)} - \beta\log \frac{\pi_{\theta}(y_l | x)}{\pi_{\text{ref}}(y_l|x)}\right)\right] 
\end{align}


where $\mathcal{D}$ is the generating distribution of the preference data ($\mathcal{D}_{X}$ is the marginal distribution of $\mathcal{D}$ on $\mathcal{X}$). Since our main goal is to study how $\mathcal{D}$ influences $\pi_{\theta}(\cdot|x)$, we take $\mathcal{D}$ as input of the loss.

\section{Properties of DPO}
\subsection{Solution for RLHF and DPO} \label{section:RLHF_sol}
In this section, we begin by introducing theoretical properties regarding DPO. For notational convenience, let us denote $f_{\theta}(x, y) = \beta\log\left(\frac{\pi_{\theta}(y|x)}{\pi_{\text{ref}}(y | x)}\right)$ 
. With this notation, DPO loss can be simplified as
\begin{align}\label{DPO_form_f}
    \mathcal{L}(\theta,\mathcal{D}) = -\mathbb{E}_{(x,y_w,y_l) \sim \mathcal{D}}[\log\sigma\left(f_{\theta}(x,y_w) - f_{\theta}(x,y_l)\right)].
\end{align}

We have our first proposition characterizing the minimizers of the RLHF and DPO loss.
\begin{proposition}\label{prop_optimal_loss}
    The function $f^*(x,y) = r^*(x,y) + c(x)$, where $c(x)$ is a function of $x$ only, is a global optimal solution to \eqref{DPO_form_f}. Consequently, the policy $\pi^*(y|x)$ defined as \begin{align}\label{RLHF_sol}
    \pi^{*}(y|x) = \frac{1}{Z(x)}\pi_{\text{ref}}(y|x)\exp\left(\frac{1}{\beta}r^*(x,y)\right),
\end{align}
where $Z(x) = \int \pi_{\text{ref}}(y|x)\exp\left(\frac{1}{\beta}r^*(x,y)\right)dy$, is optimal solution for both the RLHF formulation \eqref{RLHF_original} and the DPO formulation \eqref{DPO_BT_definition}.
\end{proposition}

We do not claim novelty for Proposition~\ref{prop_optimal_loss}, as some of its components have appeared in prior works (\cite{rafailov2023direct, azar2024general}), albeit in different forms. Rather, Proposition~\ref{prop_optimal_loss} serves to illustrate these results from a perspective that is closely aligned with our specific setup and notation.

Despite the fact that RLHF and DPO formulations share the same minimizer \eqref{RLHF_sol}, the spaces of optimal solutions \eqref{RLHF_original} and \eqref{DPO_BT_definition} are very different. This is because in DPO the policy $\pi_{\theta}$ is evaluated only on $y_w$ and $y_l$ sampled from $\mathcal{D}$. If  $supp(\mathcal{D})$ is small, it can lead to a challenging optimization landscape. Although \cite{azar2024general} briefly mentioned this phenomenon, we will give a more detailed characterization below on how $\mathcal{D}$ could affect the solution space and optimization landscape.

A policy can be optimal for the DPO loss \eqref{DPO_BT_definition} if it agrees with $\pi^*$ on $supp(\mathcal{D})$. The policy can be arbitrarily defined outside the range. To illustrate this idea more concretely, let us denote by $\mathcal{D}_{Y|x}$ the distribution of responses from $\mathcal{D}$ given a prompt $x$.

\begin{lemma} \label{lem:redefine}
Let $\pi_\theta$ be a minimizer for \eqref{DPO_BT_definition}. Then $\pi_\theta^\prime$ defined as below is also a minimizer for \eqref{DPO_BT_definition}:
\begin{align*}
    \pi_\theta^\prime(y|x) = \begin{cases} \pi_\theta(y|x) \phi(x) & \text{if } y \in supp(\mathcal{D}_{Y|x}) \\ \tilde{\pi}_\theta (y|x) & \text{otherwise} \end{cases},
\end{align*}
where $\phi$ is a function that depends only on $x$ with $\phi(x) \in$ $\left(0, 1/\int_{supp(\mathcal{D}_{Y|x})} \pi_\theta(y|x) dy\right)$, and $\tilde{\pi}_\theta$ is any policy such that $\int_{supp(\mathcal{D}_{Y|x})^c} \tilde{\pi}_\theta (y|x) dy = 1 - \phi(x)  \int_{supp(\mathcal{D}_{Y|x})} \pi_{\theta} (y|x) dy$.
\end{lemma}

This poses a possibility of a blind spot of alignment on the support of the reference policy. In LLM alignment, $\pi_{\text{ref}}$ is generally well-defined over a broad corpus, which we denote as $\mathcal{D}_{\text{ref}}$. However, DPO optimizes a model only on $supp(\mathcal{D})$, which is often constructed from human feedback. Consequently, $\mathcal{D}$ may not fully cover $supp(\mathcal{D}_{\text{ref}})$, which may lead the model to diverge on inputs that are absent in the alignment data.

\subsection{Domain-based Properties and Implications} \label{section:ref_sup}

\begin{figure}[h]
\includegraphics[width=0.4\textwidth]{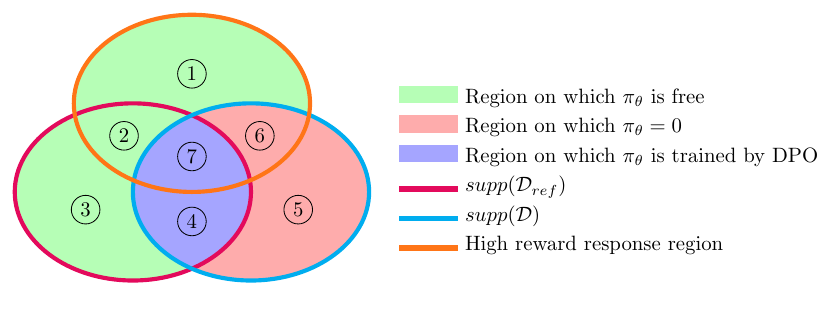}
\caption{Effect of Misaligned Supports}
\label{fig:support}
\end{figure}

\begin{tcolorbox}[colback=gray!10, colframe=gray!70, title=General Goal]
The goal of DPO alignment is to render the model to generate high-quality responses (those located in regions $1, 2, 6, 7$ of Figure~\ref{fig:support}) with higher probability after alignment.
\end{tcolorbox}

Based on the general goal of DPO, we have several insights below.

\subsubsection{The Benefit of a Good Reference Model}

Broadly speaking, the previous lemma implies that the aligned policy is heavily influenced by the reference model. Similar holds for RLHF, which we discuss in Appendix \ref{subsection:proof_supp}. We start with an observation
\begin{observation} \label{prop:DPO_supp}
Let $\pi_\theta$ be a minimizer for \eqref{DPO_BT_definition}. Then on $supp(\mathcal{D}_{Y|x})$, we have $\pi_\theta(y|x)>0$ if and only if $\pi_{\text{ref}}(y|x)>0$.
\end{observation}

\textbf{Support of the Reference Model}. The observation implies that the capability of the target policy $\pi_\theta$ is restricted by the reference model $\pi_{ref}$. This constraint can induce adverse model behavior in the context of DPO, which utilizes preference pairs and contrastive learning. Suppose that two responses $y_w$ and $y_l$ for a prompt $x$ are given in the training data. If $\pi_{ref}$ cannot generate the preferred response $y_w$, (i.e. $\pi_{ref}(y_w| x) = 0$), the target policy $\pi_\theta$ also cannot generate $y_w$ after alignment, regardless of its high quality. This corresponds to Region 6 in Figure \ref{fig:support}.

\textbf{High Probability on High-reward Region}.  It is recommended that high-quality responses also have a non-negligible probability under the reference model. As shown in Figure \ref{fig:ref_well}-\ref{fig:training_curve} in Appendix \ref{subsection:ref_model_impact}, a policy will be much less capable of producing desirable responses if $\pi_{ref}$ assigns these responses a low probability. This can be achieved either through \textbf{careful data curation} or by selecting a \textbf{stronger reference model}. Choosing a reference model that can generate good responses expands Region 7, further helping the model training. 

We also note that isolating the effect of the reference model on alignment is challenging. First, with modern complex LLMs, any changes to the implicit response distribution are inherently difficult to control. Moreover, models from the same family (e.g., LLaMA 3.1) exhibit minimal performance differences, while models from different architectures introduce confounding factors that render direct comparisons uninformative. To address this, we use our proposed linear alignment model (Section~\ref{section:linear}), which offers greater variability in a controlled setting and enables theoretical analysis.



\subsubsection{The Benefit of High-quality Alignment Dataset }

Lemma \ref{lem:redefine} and Figure \ref{fig:support} imply several limitations of the current DPO training pipeline. In practice, the DPO loss is evaluated solely on $\mathcal{D}$, which is often collected manually. $\pi_{\text{ref}}$, on the other hand, typically corresponds to the policy obtained after supervised fine-tuning (SFT) in the post-training phase, which is independent of $\mathcal{D}$. This will cause the following issues.
    \begin{itemize}
    \item \textbf{Flat plateau for optimization}. There is no guarantee that $supp(\mathcal{D})$ covers $supp(\mathcal{D}_{\text{ref}})$. The resulting optimization problem inherently suffers from a flat plateau on $supp(\mathcal{D}_{\text{ref}})\cap supp(\mathcal{D})^c$ (regions 2 and 3 in Figure \ref{fig:support}) in the optimization domain, presenting fundamental challenges for effective optimization.
    \item \textbf{Incomplete coverage and zero probability for region of good responses}. If the training data or the reference policy is suboptimal, $supp(\mathcal{D})$ and $supp(\mathcal{D}_{\text{ref}})$ might fail to include many of the responses with high reward.
    Consequently, the target policy does not receive meaningful learning signals on regions 1 and 2, where the loss is not evaluated. In addition, the target policy will be 0 in region 6.
    \end{itemize}
    
The issue above can be addressed by improving the quality of the dataset $\mathcal{D}$ and expanding its coverage. This increases Regions 1, 2, 3, and 6, directly mitigating the problem.

\subsubsection{Adopting Online/Iterative DPO Training} 
Our discussion above also suggests the advantage of multi-step online/iterative training. In a multi-round setup, we can update and improve both the reference policy $\pi_{\text{ref}}$ and the generating distribution $\mathcal{D}$ in subsequent rounds. This will shift the range of the trained policy toward areas that may contain high-reward responses, making the optimization problem less challenging. For generality, we provide a standard online DPO pipeline in Algorithm \ref{alg:online_DPO}, which will be discussed later. Please note that we provide the algorithm description just for reference.

\begin{algorithm}[ht!]
\caption{Online DPO ($\theta, T, \mathcal{D}, \text{ALG}_{\text{Data}}$)}
\label{alg:online_DPO}
\begin{algorithmic}[1]
\State Input: parameter $\theta\in\Theta$ for an LLM; number of iterations $T$; initial (human preference) data $\mathcal{D} = \{(x^{(i)},y_w^{(i)},y_l^{(i)})\}_{i=1}^N$; data generating algorithm $\text{ALG}_{\text{Data}}$
\State Set $t = 0$, $\theta_t = \theta$, and $\mathcal{D}^{(t)} = \mathcal{D}$.
\For{$t < T$}
\State $\theta_{t+1} = \argmin_{\theta\in\Theta} \mathcal{L}(\theta,\mathcal{D}^{(t)})$ defined in \eqref{DPO_BT_definition}, with $\theta_{\text{ref}} = \theta_t$
\State (Optional) Update $\theta_{t+1}$ using other post-training algorithm (e.g. fine-tuning)
\State Generate $\mathcal{D}_{t+1}$ using $\text{ALG}_{\text{Data}} (\theta_{t+1}, \mathcal{D})$, and update $t = t+1$
\EndFor
\State Output: $\theta_{T}$
\end{algorithmic}
\end{algorithm}

\subsection{Gradient-based Properties and Implications} \label{section:gradient}
In this section, we present results characterizing how response quality influences one-step gradient descent updates in DPO. Following the theorem, we draw connections to the well-known phenomenon of likelihood displacement, as studied in prior works \citep{shi2024understanding, razin2024unintentional}. We start with a comprehensive definition of the data generation process for $\mathcal{D}$.

\textbf{Data Generation Process}. Throughout the paper, the sampling mechanism $(x,y_w,y_l) \sim \mathcal{D}$ follows
\begin{itemize}
    \item First, a prompt $x$ is sampled from a distribution $\mathcal{D}_X$ which admits a pmf (or density) $p_X(\cdot)$
    \item For the generated prompt $x$, sample two \textbf{ordered} responses $y$ and $y'$ from a specified distribution $\mathcal{D}_{Y_1,Y_2|X=x}$ with pmf or density $p_{Y_1,Y_2|x}(\cdot,\cdot|x)$. Notice that $y$ and $y'$ need not be i.i.d., and the responses could either be human-generated or AI-generated. 
    \item To facilitate theoretical analysis, we denote by $\mathcal{D}_u$ the generating distribution for \textbf{unordered} tuple from $\mathcal{D}_{Y_1,Y_2|X=x}$. Equivalently, we sample $(x, y, y') \sim \mathcal{D}_u$ and treat $(x, y, y')$ and $(x, y', y)$ as the same. We denote by $q_{Y_1,Y_2|x}(\cdot,\cdot)$ the corresponding unordered pmf.
    \item Then based on the sampled $y$ and $y'$, a BT model is applied to label $y_w$ and $y_l$ via $\mathbb{P}(y = y_w|x) = \mathbb{P}(y \succ y'|x)= \sigma(r^*(x,y) - r^*(x,y'))$. We denote by $p_{Y_w,Y_l|x}(\cdot,\cdot|x)$ the resulting pmf (or density) of $y_w$ and $y_l$. Note that $p_{Y_W,Y_l | x}(y,y'|x)$ is not necessarily the same as $p_{Y_W,Y_l | x}(y',y|x)$, and thus the order of the two inputs of $p_{Y_W,Y_l | x}(\cdot,\cdot | x)$ does matter. 
\end{itemize}

To quantify model improvement, we fix a prompt $x$ and consider responses $y$ with high reward $r^*(x, y)$. Ideally, a well-trained policy $\pi_\theta(\cdot|x)$ should assign higher likelihood to such responses. The following theorem characterizes when $\pi_\theta(y|x)$ increases or decreases after a gradient step. It depends on two key quantities. First, the ``true winning probability'' of $y$, conditioned on it appearing in a preference pair, is defined as
\begin{align*}
    \mathbb{P}_w(y|x) &= \mathbb{E}_{Y_2 \sim \mathcal{D}_u | (X,Y_1)=(x,y)} \left[\mathbb{P}(y \succ Y_2 \mid x)\right] \\
    &= \int_{\mathcal{D}_Y} \mathbb{P}(y \succ y_2 \mid x) \, 2q_{Y_1,Y_2|x}(y,y_2 \mid x) dy_2.
\end{align*}
Next, the ``model-implied winning probability'' under $\theta$ is defined as
\begin{align*}
    \mathbb{P}_{w,\theta}(y|x) &= \mathbb{E}_{Y_2 \sim \mathcal{D}_u | (X,Y_1)=(x,y)} \left[\mathbb{P}_\theta(y \succ Y_2 \mid x)\right] \\
    &= \int_{\mathcal{D}_Y} \mathbb{P}_\theta(y \succ y_2 \mid x) \, 2q_{Y_1,Y_2|x}(y,y_2 \mid x) dy_2,
\end{align*}
with $\mathbb{P}_\theta(y_1 \succ y_2 \mid x) = \sigma(f_\theta(x, y_1) - f_\theta(x, y_2))$ being the winning probability implied by the current model parameter $\theta$.

\begin{theorem}\label{thm:gradiennt}
Let $\pi_{\theta_{t+1}}$ be the policy after one step of gradient descent on $\mathcal{L}(\theta_t,\mathcal{D})$ defined in \eqref{DPO_BT_definition} at step $t$. Then the updated policy is compared with the previous policy as 
\begin{enumerate}[i)]

\item $\pi_{\theta_{t+1}}(y|x) >  \pi_{\theta_t}(y|x) \,\,\,\text{if } ~ \mathbb{P}_w(y|x) > \mathbb{P}_{w,\theta_t}(y|x)$
\item $ \pi_{\theta_{t+1}}(y|x) =  \pi_{\theta_t}(y|x) \,\,\,\text{if } ~ \mathbb{P}_w(y|x) = \mathbb{P}_{w,\theta_t}(y|x)$ or $y \notin \text{supp}(\mathcal{D}_{Y|x})$
\item $\pi_{\theta_{t+1}}(y|x) <  \pi_{\theta_t}(y|x) \,\,\,\text{if } ~ \mathbb{P}_w(y|x) < \mathbb{P}_{w,\theta_t}(y|x)$.
\end{enumerate}
Moreover, the change in the log-probability $f_{\theta_{t+1}}(y) - f_{\theta_t}(y)$ is proportional to the probability gap $\mathbb{P}_w(y|x) - \mathbb{P}_{w,\theta_t}(y|x)$ up to $O(\alpha^2)$ by 
\begin{align}\label{eqn:gradients}
    f_{\theta_{t+1}} (x,y) & - f_{\theta_t}(x,y) = 2\alpha  \left(\mathbb{P}_w(y|x) - \mathbb{P}_{w,\theta_t}(y|x)\right) \nonumber \\
    &\times p_{X,Y_1}(x,y) \left.\frac{\partial f_\theta}{\partial \theta}\right\vert_{\theta=\theta_t}(x,y)^2 + O(\alpha^2).
\end{align}
\end{theorem}
The theorem implies several aspects that we want to highlight.
\begin{itemize}
    \item \textbf{True vs. Implied BT Probability.} DPO increases the likelihood of a response $y$ when its \textit{true winning probability} under the Bradley–Terry (BT) model exceeds the \textit{model-implied winning probability} under the current parameters $\theta$. Specifically, DPO updates the policy $\pi_{\theta_t}$ by comparing the model’s estimate with the BT ground truth: if $\mathbb{P}_w(y|x) > \mathbb{P}_{w,\theta_t}(y|x)$, then DPO increases $\pi_{\theta_{t+1}}(y|x)$ accordingly.
    
    \item \textbf{Probability Gap as Gradient Signal.} A large gap between $\mathbb{P}_w(y|x)$ and $\mathbb{P}_{w,\theta_t}(y|x)$ indicates that the model underestimates the true preference for $y$. In such cases, DPO applies a larger update to $\pi_\theta(y|x)$. Thus, the magnitude of the update reflects the discrepancy between the model’s prediction and the underlying preference signal.
    
    \item \textbf{Effect of $\text{supp}(\mathcal{D})$.} As shown in the bullet point above and \eqref{eqn:gradients}, if a response $y$ never appears in $\mathcal{D}$, then no preference signal is available for it, and DPO does not update the policy at $y$. Consequently, if the initial policy performs poorly on $y \in \text{supp}(\mathcal{D})^c$, its performance will remain poor after training, since $\pi_\theta(y|x)$ remains unchanged.
\end{itemize}

\subsection{Semantic Correlation}

\begin{figure}
    \centering
    \includegraphics[width=\linewidth]{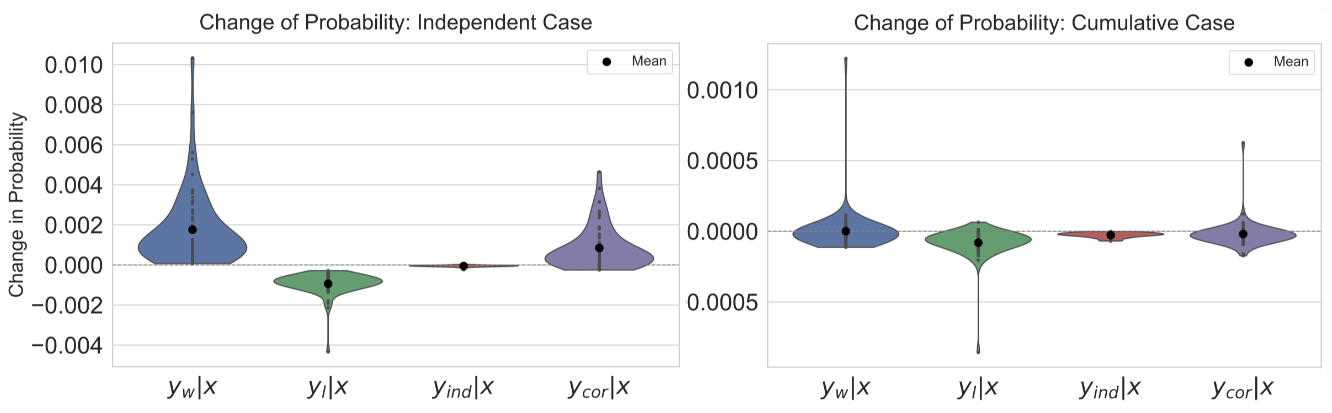}
    \caption{$y_{\text{ind}}$ refers to the response that is not semantically related to $y_w$ or $y_l$, and $y_{\text{cor}}$ refers to the response semantically correlated with $y_w$. The left diagram shows the change of probability after applying one step of gradient descent on the entire batch of data set $\{(x^{(i)},y_w^{(i)},y_l^{(i)})\}_{i=1}^N$. The right diagram shows the change of probability after applying one step of gradient descent on the entire batch of data set $\{(x^{(i)},y_w^{(i)},y_l^{(i)})\}_{i=1}^N$.}
    \label{fig:displacement}
\end{figure}

To numerically present the insights of Theorem \ref{thm:gradiennt}, we begin with an experiment in which the dataset $\mathcal{D}$ contains only a single tuple ${(x, y_w, y_l)}$. We refer to this as the independent case, as no other response pair directly influences the outcome. After performing one step of gradient descent on $\mathcal{L}(\theta, \mathcal{D})$, we observe that the updated policy $\pi_{\theta'}$ assigns a higher probability to $y_w$ and a lower probability to $y_l$. For other responses $y \notin {y_w, y_l}$, the change in probability $\pi_{\theta'}(y|x)$ remains close to zero, as illustrated in Figure~\ref{fig:displacement}. These patterns are consistent with the predictions of Theorem~\ref{thm:gradiennt}.

However, an important nuance is that responses often exhibit strong \textbf{semantic correlations}. In particular, we observe a noticeable increase in the probability of $y_{\text{cor}}$, a response that is semantically similar to $y_w$. While this change is not directly mandated by the update rule, it is consistent with the behavior one would expect given the structure of the model's underlying language representations.
Such semantic correlations are largely harmless when the dataset contains only a single tuple. However, as we scale up the size of $\mathcal{D}$, we encounter a well-known practical issue called \textbf{the likelihood displacement}, using a dataset of $N$ preference pairs $\mathcal{D} = \{(x^{(i)}, y_w^{(i)}, y_l^{(i)})\}_{i=1}^N$. Likelihood displacement refers to the counterintuitive phenomenon where the likelihood of both preferred and dispreferred responses decreases after DPO training \citep{shi2024understanding, razin2024unintentional}. In the right panel of  Figure~\ref{fig:displacement}, the average change in probability for the chosen response is slightly below 0, unexpectedly. Such a phenomenon complicates the verification of theoretical results, including in our setting. In principle, the gradient contribution from each individual tuple should align with the intended behavior, as shown in the left panel of Figure~\ref{fig:displacement}. Yet, due to semantic correlations, the gradient update for one response can inadvertently influence semantically related responses. Consequently, when aggregated over the dataset, the gradient may drift in an unintended direction, thereby diminishing the intended learning effect on $y_w$ and $y_l$. Note that, unlike the finding in \citet{razin2024unintentional}, where the correlation arises directly from the tokens within the preference pair $(y_w, y_l)$, our finding reveals that other preference pairs within the same batch can also influence the likelihood of $(y_w, y_l)$. Additional experimental details are provided in Appendix~\ref{app:exp}.

Therefore, the key takeaway is that we need a model capable of mitigating the unexpected effects of the intractable semantic correlations, while still preserving the core of DPO for understanding new policies.

\section{Mitigating Semantic Correlations: Online Linear Alignment Model} \label{section:linear}

 



In the previous sections, our analyses of DPO training dynamics face several sources of intractability, including the dependence on $\mathcal{D}$, $\pi_{\text{ref}}$, the corresponding support, and the semantic correlation among data. 
In order to better focus on the influence of the higher-quality responses $y_w$ on the training dynamics of DPO, we consider a simplified setting: a linear model with a Gaussian-based policy. The Gaussian policy ensures tractability for both the RLHF and DPO settings; the prompt and responses are no longer sequences, avoiding semantic correlation. 
Based on this framework, we will first theoretically prove that the standard online DPO algorithm can converge to the true model. Moreover, we show that improving the quality of $\mathcal{D}$ can accelerate the convergence of DPO and enhance training efficiency. 

\paragraph{Linear Alignment Model Setup.} For every input-response pair $(x, y)$, we assume a target model $y = w^{* \top} x$ for some $w^* \in \mathbb{R}^d$. For analytical tractability, we take the reward to be $
    r^*(x, y) = -||w^{* \top}x - y||^2_2.$
Our goal is to find a Gaussian policy $\pi_{\theta}$, parameterized by $\theta = (w,\sigma)$, such that $y = w^{\top} \, x + \sigma  \epsilon$ for a given $x$ where $\epsilon\sim\mathcal{N}(0,1)$. The policy
is expressed as $\pi_{\theta}(\cdot|x) \sim \mathcal{N}(w^\top x, \sigma)$, or $
    \pi_{\theta}(y|x) = \frac{1}{\sqrt{2\pi \sigma^2}}\exp\left(-\frac{(y-w^\top x)^2}{2\sigma^2}\right)$.
We have the two properties, \textit{free of likelihood displacement} and \textit{theoretical convergence}, for our linear alignment model.

\begin{figure}[t]
    \centering
    \includegraphics[width=\linewidth]{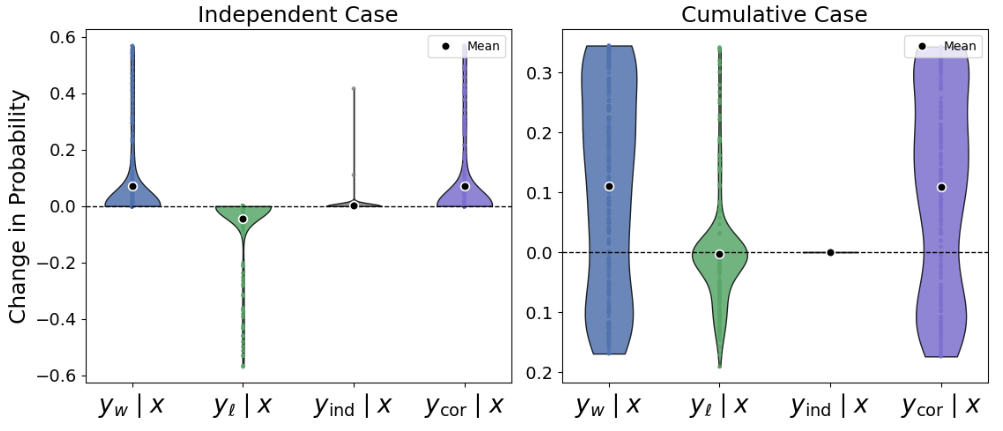}
    \caption{$y_{\text{ind}}$ refers to the response that is significantly different from to $y_w$ or $y_l$, and $y_{\text{cor}}$ refers to the vector close to $y_w$. The left diagram shows the change of probability after repeatedly applying one step of gradient descent on one tuple $(x,y_w,y_l)$ from the entire dataset. The right diagram shows the change of probability after applying one step of gradient descent on the entire batch of data set $\{(x^{(i)},y_w^{(i)},y_l^{(i)})\}_{i=1}^N$.}
    \label{fig:displacement_linear}
\end{figure}

\textbf{Free of Likelihood Displacement.} In Figure~\ref{fig:displacement_linear}, we replicate the experiment from Figure~\ref{fig:displacement} using our linear alignment model. The left panel shows that training on a single data point behaves reliably, confirming the effectiveness of our model at the level of an individual tuple. Moreover, when we scale up the dataset size (right panel), the average change in probability remains positive, in stark contrast to Figure~\ref{fig:displacement}. While a noticeable portion of samples still fall below zero, the overall trend is positive. This demonstrates that our linear model effectively avoids likelihood displacement, providing a reliable proxy for analysis.

\textbf{Convergence.} We now present the theoretical convergence result based on the linear alignment model.
\begin{lemma}\label{lem:linear_RLHF}
    Starting at some reference policy $\pi_{\theta_{\text{ref}}}$ with $\theta_{\text{ref}} = (w_{\text{ref}}, \sigma_{\text{ref}})$, the policy $\pi_{\theta^*}$ that minimizes RLHF loss \eqref{RLHF_original} follows
    \begin{align}\label{eqn_RLHF_LS_sol}
        \pi_{\theta^*}(\cdot|x) \sim \mathcal{N}\left( (\gamma w_{\text{ref}} + (1-\gamma)w^*)^\top x\,, \,\,\frac{\sigma_{\text{ref}}^2\beta}{\beta+2\sigma_{\text{ref}}^2}\right), 
\end{align}
where $\gamma =\frac{\beta}{\beta+2\sigma_{\text{ref}}^2}$.
\end{lemma}

Lemma \ref{lem:linear_RLHF} states that the policy minimizing the RHLF loss has a mean of the linear combination $\gamma w_{\text{ref}} + (1 - \gamma) w^*$. This form suggests that if we take the reference policy to be the minimized policy in the previous iteration and repeat this process (see Step 4 in Algorithm \ref{alg:online_DPO}), the trained policy will converge to $w^{*\top}x$.

 Lemma~\ref{lem:linear_RLHF} also facilitates the analysis of the convergence behavior of the online DPO framework in Algorithm~\ref{alg:online_DPO}, under the following assumption that addresses the identifiability and optimization issues discussed earlier.

\begin{assumption} \label{assumption:linear_RLHF}
    The generating distributions $\{\mathcal{D}^{(t)}\}_{t=1}^T$ have full support in terms of $y$.
\end{assumption}
\begin{proposition} [Convergence of Online OPO]\label{prop_linear_RLHF}
Consider the linear alignment model, and apply Algorithm~\ref{alg:online_DPO} with $\theta_0 = (w_0, \sigma_0)$, number of iterations $T$, and data generation procedure $\text{ALG}_{\text{Data}}$ satisfying Assumption~\ref{assumption:linear_RLHF}. At iteration $t$, the minimizing policy $ \pi_{\theta_t}(\cdot \mid x) \sim \mathcal{N}(w_t^\top x, \sigma_t^2)$, where
    \begin{align*}
        w_t= w^* + \frac{\beta}{\beta+2t\sigma_0^2}(w_0-w^*) \quad \text{and} \quad \sigma_t^2 = \frac{\beta \sigma_0^2}{\beta + 2t \sigma_0^2}.
    \end{align*}
    Moreover, the generated response $Y_t \sim \pi_{\theta_t}(\cdot \mid x)$ converges to $w^{*\top} x$ in probability as $t \to \infty$.
\end{proposition}

\subsection{Generating Distribution on Learning Dynamics} \label{section:convergence}
Theorem~\ref{thm:gradiennt} establishes the theoretical foundation for studying the influence of the data-generating distribution. We take standard sampling as the benchmark and use Best-of-$K$ sampling as an illustrative example to improve the quality of chosen responses. The advantage of Best-of-$K$ is straightforward: by easily increasing $K$, one can reliably obtain higher-quality responses. Importantly, Best-of-$K$ is not the only applicable scheme: other samplers proposed in \citet{shi2024crucial,feng2025pilaf,xiong2024iterative,tajwar2024preference,guo2024direct} can also be incorporated, and our analysis continues to hold under these alternatives.

We now consider two general data-generating scenarios: the standard sampling as a benchmark and the Best-of-$K$ sampling to improve the chosen responses. We want to maintain a minimalist yet general design, and it is likely that other sampling variations would also be effective. See Algorithm \ref{alg:Data} for a detailed implementation.

\vspace{-2mm}
\begin{algorithm}[ht!]
\caption{$\text{ALG}_{\text{Data}} (\theta, \mathcal{D}, I)$}
\label{alg:Data}
\begin{algorithmic}[1]
\State Input: parameter $\theta\in\Theta$ of an LLM; dataset $\mathcal{D}$; $I$, indicator for the sampling mode ($1$ or $2$)
\State Initialize dataset $\mathcal{D}_{gen} = \{\}$
\For {$x \in \mathcal{D}_X$}
\If{$I == 1$} \label{alg_line:sampling_start}
\State \textbf{Case 1 (Standard Sampling)} Generate $y_1, y_2$ independently from $\pi_{\theta}(\cdot|x)$
\Else{ \,($I == 2$)}
\State \textbf{Case 2 (Best-of-$K$ Sampling)} 
Sample $K$ responses $\{y^{(k)}\}_{k=1}^K$ and $y_2$ all independently.
\State Take $y_1 = \argmax_{\,y^{(k)}\in \{y^{(k)}\}_{k=1}^K} r^*(x,y^{(k)})$.
\EndIf \label{alg_line:sampling_end}
\State Based on $y_1, y_2$, sample $y_w, y_l$ with $\mathbb{P}(y_w=y_1 \succ y_l=y_2|x) = \sigma(r^*(x,y_1) - r^*(x,y_2))$
\State $\mathcal{D}_{gen} = \mathcal{D}_{gen}\cup (x,y_w,y_l)$
\EndFor
\State Return $\mathcal{D}_{gen}$
\end{algorithmic}
\end{algorithm}

To establish theoretical results, we first define several key quantities. For the current parameter $(w, \sigma)$ and a fixed input $x$, define the bias as $\delta(x) = ((w - w^*)^{\top} x)/\sigma$. Let $\epsilon_1 = (y_1 - w^{\top} x)/\sigma$ denote the standardized noise associated with $y_1$ (as defined in Steps \ref{alg_line:sampling_start}–\ref{alg_line:sampling_end} of Algorithm~\ref{alg:Data}), and similarly, let $\epsilon_2$ denote the standardized noise associated with $y_2$.

Additionally, as $\mathbb{E}[\epsilon_1^2]$ and $\mathbb{E}[|\epsilon_1 - \epsilon_2|]$ play a key role in the theoretical results, we give their analytical forms. These quantities depend on the bias $\delta(x)$ and the sampling parameter $K$.

\begin{lemma}\label{lem:analytical}
    For a given $x$, $\mathbb{E}[\epsilon_1^2]$ and $\mathbb{E}[|\epsilon_1 - \epsilon_2|]$ are functions of $K$ and $\delta(x)$ such that 
    \begin{align*}
        \eta(K, \delta(x)) &:= \mathbb{E}[\epsilon_1^2] \\
        &= K\int z^2\varphi(z) \left(1-F_{|\delta(x)+Z|}(|\delta(x)+z|)\right)^{K-1}dz,\\
        \gamma(K, \delta(x)) &:= \mathbb{E}[|\epsilon_1-\epsilon_2|] \\
        &= K\int \varphi(z) \left(1-F_{|\delta(x)+Z|}(|\delta(x)+z|)\right)^{K-1} \\
        &\times \left(z(2\Phi(z) - 1) + 2\varphi(z) \right)dz,
    \end{align*}
    where $\varphi$ and $\Phi$ denote the standard normal density and CDF, respectively, and $F_{|\delta(x)+Z|}(\cdot)$ is the CDF of the random variable $|\delta(x) + Z|$ with $Z \sim \mathcal{N}(0,1)$.
\end{lemma}
Lemma~\ref{lem:analytical} serves as a foundation for the following proposition, allowing us to rigorously examine the curvature and gradient magnitude, two fundamental quantities for capturing the training dynamics of DPO.
\begin{proposition}[First and second order properties]\label{prop_fisher}
Let $N$ denote the sample size of $\mathcal{D}$, with $\{x_n\}_{n=1}^N$ representing the current sampled inputs, and let $t$ denote the current iteration of the online DPO training. For the current parameter $w_t$, the magnitude of the gradient is bounded by
\begin{align*}
\lVert \nabla_w\mathcal{L}(w_t) \rVert \leq \begin{cases}
    \frac{\beta}{2N\sigma_{t}}\sum_{n=1}^N \gamma(K,\delta(x_n)) \Vert x_n \rVert  & K > 1\\
    \frac{\beta}{2N\sigma_{t}} \frac{1}{\sqrt{\pi}}\sum_{n=1}^N  \Vert x_n \rVert  & K = 1
\end{cases} 
\end{align*}
and the Fisher information matrix is
\begin{align*}
I_K(w_t) = \begin{cases}
    \frac{\beta^2}{4N\sigma_{t}^2} \sum_{n=1}^N \left\{ \eta(K, \delta(x_n)) +1 \right\} x_n x_n^\top & K > 1\\
    \frac{\beta^2}{2N\sigma_{t}^2} \sum_{n=1}^N x_n x_n^\top  & K = 1
\end{cases}
\end{align*}
\end{proposition}
\begin{figure}[t]
    \centering
    \includegraphics[width=0.36\textwidth]{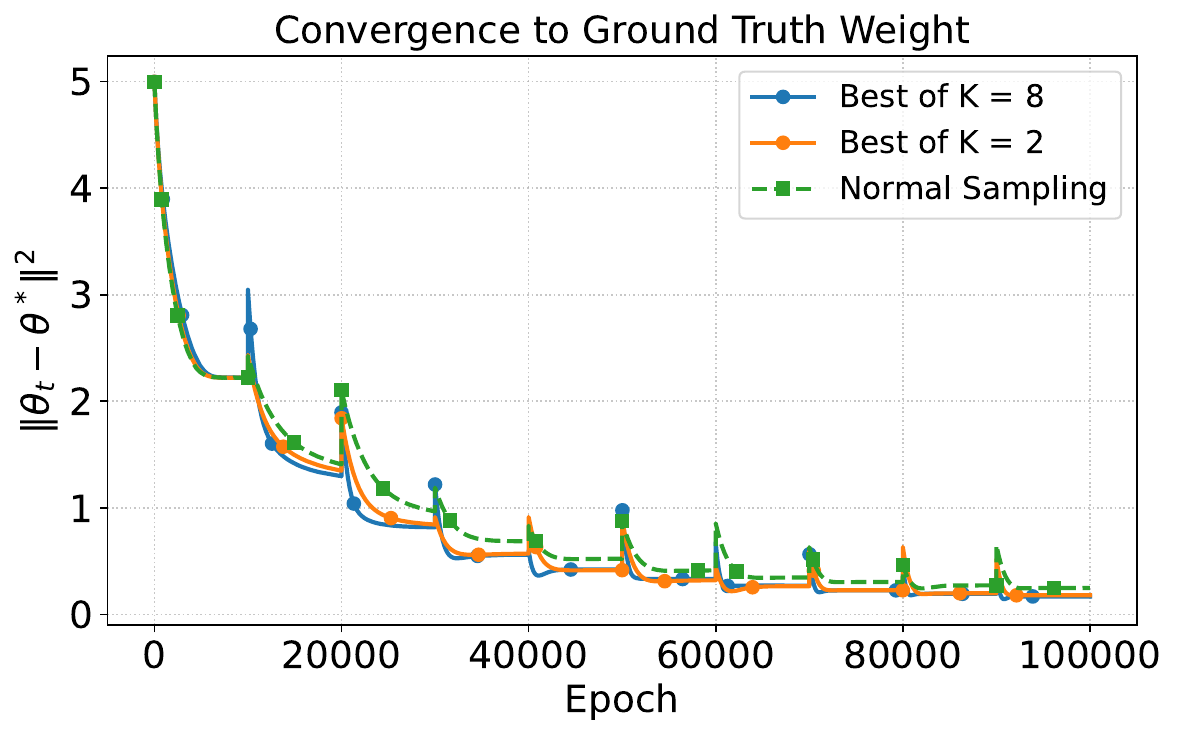}
    \caption{Convergence of Algorithm~\ref{alg:Data} to the ground-truth weight in the linear alignment model.}
    \label{fig:gradient}
    \vspace{-5mm}
\end{figure}

\vspace{-3mm}
Larger values of curvature (approximated by the Fisher information in the DPO setting) and greater gradient magnitudes generally help optimization. The above proposition suggests that Best-of-$K$ sampling is effective when $\gamma(K, \delta(x_n)) > \frac{1}{\sqrt{\pi}}$ and $\eta(K, \delta(x_n)) > 1$. Limit approximations (detailed in Appendix \ref{section:proof}) show that $\gamma(K, \delta(x)) \approx |\delta(x)|$ and $\eta(K, \delta(x)) \approx \delta(x)^2$ for large $|\delta(x)|$, while for small $|\delta(x)|$, $\gamma(K, \delta(x)) \geq \sqrt{2/\pi}$ and $\eta(K, \delta(x)) \leq 0.5$. These results underscore the advantage of Best-of-$K$ sampling in enhancing both gradient strength and curvature, particularly when the bias $\delta(x)$ is large. See Figure~\ref{fig:gradient} for empirical evidence for the best-of-$K$ sampling. 

\vspace{-3mm}

\section{Numerical Experiment} \label{section:experiment}
\vspace{-3mm}
In this section, we present a series of experiments on LLMs following the procedures in Algorithms~\ref{alg:online_DPO} and \ref{alg:Data}. By generating multiple responses using an LLM and selecting the one with the highest reward according to a given reward model, we observe consistent improvements across iterations. Furthermore, our model trained via online DPO show no sign of reward hacking in the responses generated by the trained DPO policy using AI-improved feedback. Due to space limitations, we refer readers to Appendix~\ref{app:exp} for further details.

\textbf{Experiment Setup}. In our experiment, we adopt \texttt{Llama-3.1-Tulu-3-8B-SFT} \citep{lambert2024tulu3} as the reference policy and use \texttt{Skywork-Reward-Llama-3.1-8B-v0.2} \citep{liu2024skywork} as the reward model. The initial preference dataset is from UltraFeedback \citep{cui2024ultrafeedbackboostinglanguagemodels}. We repeatedly improve the preferred responses to enhance the training signal and evaluate model performance on both in- and out-of-distribution prompts. See Appendix~\ref{section:experiment_setting1} for full details.


We compare the following three configurations for DPO training:
\vspace{-3.9mm}
\begin{itemize}
\item $\textbf{DPO}_{\text{base}}$: Starting at $\theta_{\text{ref}}$ and perform DPO with the original preference dataset $\mathcal{D}$.

\item $\textbf{DPO}^+$: We improve the responses in $y_w$ and generate $\mathcal{D}^+$ using the responses sampled by the reference model and select the top response using the reward model. Then, we  start at $\theta_{\text{ref}}$ and perform DPO with the improved preference dataset $\mathcal{D}^+$.

\item $\textbf{DPO}^{++}$: We improve the responses in $y_w$ and generate $\mathcal{D}^{++}$ using the responses sampled by $\text{DPO}^+$, and select the best one according to the reward model. Then, we start at $\theta_{\text{DPO}^+}$ and perform DPO with the dataset $\mathcal{D}^{++}$.
\end{itemize}
\vspace{-3mm}

\begin{figure}[t]
    \centering
    \includegraphics[width=0.8\linewidth]{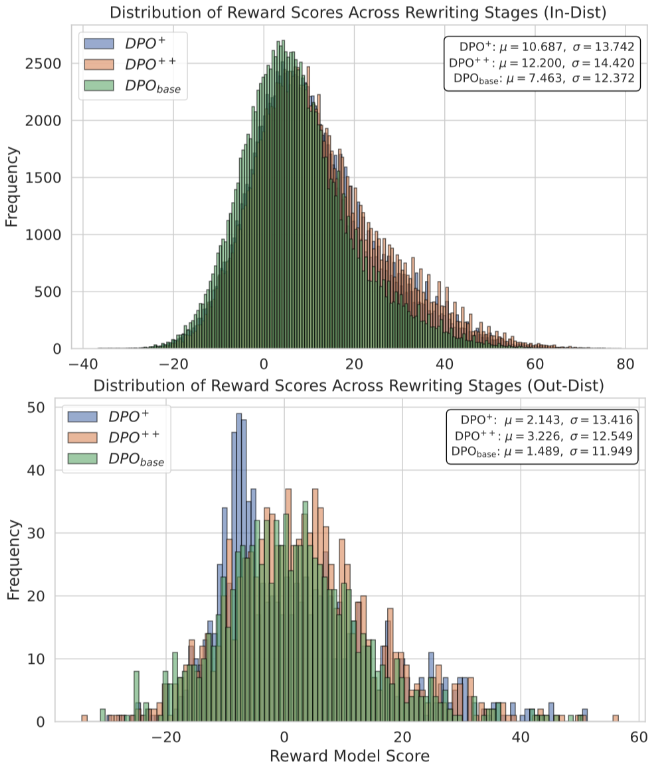}
    \caption{Sampled responses' reward distribution for different models (ID and OOD).}
    \label{fig:indist_distribution}
    \label{fig:ood_distribution}
\end{figure}



\begin{minipage}{\linewidth}
\centering
\captionof{table}{Pairwise Win Rates of Different Models}
\begin{tabular}{@{}lc@{}}
\toprule
\textbf{Comparison} & \textbf{Win Rate (\%)} \\
\midrule
$DPO^{+}$ vs. $DPO_{base}$ & 52.7 \\
$DPO^{++}$ vs. $DPO^{+}$ & 52.1 \\
$DPO^{++}$ vs. $DPO_{base}$ & 58.6 \\
\bottomrule
\end{tabular}
 \label{tab:rewrite_winrate}
\end{minipage}


Figures~\ref{fig:indist_distribution} demonstrate that improving the quality of preference data enhances DPO training and results in policies generating higher-reward responses. With more data-rewriting stages, the reward of the responses in both ID and OOD settings increases at a distributional level. We also test the win rate of these models on the validation set in Table \ref{tab:rewrite_winrate}, which aligns with our previous findings.

\vspace{5.7mm}
\section{Conclusion}
We investigate the role of the sampling distribution in DPO and its influence on training dynamics. Through both theoretical analysis and empirical validation, we demonstrate that the support and quality of the generating distribution play a crucial role in shaping the solution space and convergence behavior of DPO. Our gradient-based analysis highlights how the distribution of responses drives policy updates, offering insights into common empirical observations. By introducing a simplified alignment model that is free of likelihood displacement, we quantify how frequent high-quality responses amplify the gradient signal, enhancing the optimization landscape and accelerating effective learning. These results offer theoretical grounding for practical strategies such as online and adaptive DPO, emphasizing the importance of careful data design and sampling in preference-based training.


\bibliography{sample_paper}


\section*{Checklist}

\textcolor{green}{From the authors: our responses are marked in green.}

\begin{enumerate}

  \item For all models and algorithms presented, check if you include:
  \begin{enumerate}
    \item A clear description of the mathematical setting, assumptions, algorithm, and/or model. [\textcolor{green}{Yes}/No/Not Applicable]
    \item An analysis of the properties and complexity (time, space, sample size) of any algorithm. [Yes/No/\textcolor{green}{Not Applicable}]
    \item (Optional) Anonymized source code, with specification of all dependencies, including external libraries. [Yes/No/Not Applicable]
  \end{enumerate}

  \item For any theoretical claim, check if you include:
  \begin{enumerate}
    \item Statements of the full set of assumptions of all theoretical results. [\textcolor{green}{Yes}/No/Not Applicable]
    \item Complete proofs of all theoretical results. [\textcolor{green}{Yes}/No/Not Applicable]
    \item Clear explanations of any assumptions. [\textcolor{green}{Yes}/No/Not Applicable]     
  \end{enumerate}

  \item For all figures and tables that present empirical results, check if you include:
  \begin{enumerate}
    \item The code, data, and instructions needed to reproduce the main experimental results (either in the supplemental material or as a URL). [\textcolor{green}{Yes}/No/Not Applicable]
    \item All the training details (e.g., data splits, hyperparameters, how they were chosen). [\textcolor{green}{Yes}/No/Not Applicable]
    \item A clear definition of the specific measure or statistics and error bars (e.g., with respect to the random seed after running experiments multiple times). [\textcolor{green}{Yes}/No/Not Applicable]
    \item A description of the computing infrastructure used. (e.g., type of GPUs, internal cluster, or cloud provider). [\textcolor{green}{Yes}/No/Not Applicable]
  \end{enumerate}

  \item If you are using existing assets (e.g., code, data, models) or curating/releasing new assets, check if you include:
  \begin{enumerate}
    \item Citations of the creator If your work uses existing assets. [\textcolor{green}{Yes}/No/Not Applicable]
    \item The license information of the assets, if applicable. [Yes/No/\textcolor{green}{Not Applicable}]
    \item New assets either in the supplemental material or as a URL, if applicable. [Yes/No/\textcolor{green}{Not Applicable}]
    \item Information about consent from data providers/curators. [\textcolor{green}{Yes}/No/Not Applicable]
    \item Discussion of sensible content if applicable, e.g., personally identifiable information or offensive content. [Yes/No/\textcolor{green}{Not Applicable}]
  \end{enumerate}

  \item If you used crowdsourcing or conducted research with human subjects, check if you include:
  \begin{enumerate}
    \item The full text of instructions given to participants and screenshots. [Yes/No/\textcolor{green}{Not Applicable}]
    \item Descriptions of potential participant risks, with links to Institutional Review Board (IRB) approvals if applicable. [Yes/No/\textcolor{green}{Not Applicable}]
    \item The estimated hourly wage paid to participants and the total amount spent on participant compensation. [Yes/No/\textcolor{green}{Not Applicable}]
  \end{enumerate}
\end{enumerate}

\clearpage
\appendix
\thispagestyle{empty}

\onecolumn
\aistatstitle{Appendix}

\section{Related Work}\label{app:related_work}

\paragraph{Reinforcement Learning and Alignment of Large Language Models} 
Pretrained and fine-tuned LLMs are further aligned to provide helpful, honest, and harmless responses. \cite{stiennon2020learning} aligned LLMs and demonstrated that alignment improves the quality of responses. \cite{ouyang2022training} proposed the framework of reinforcement learning with human feedback(RLHF), which trains a reward model to rank responses and a policy to maximize the reward. \cite{bai2022training} also employed RLHF and showed performance improvement across assorted NLP tasks. However, training a reward model and a policy requires substantial engineering effort, including TRPO \citep{schulman2015trust} and PPO \citep{schulman2017proximal}, pointed out by \cite{choshen2020on}. 

Instead of training a separate reward model, a policy can be directly optimized. \cite{rafailov2023direct} proposed direct preference optimization (DPO), a method that eliminates the need for a separate reward model by directly optimizing the policy. This is achieved by leveraging the closed-form solution to the RLHF objective, under the assumption that true preferences follow the Bradley-Terry model introduced by \cite{bradley1952rank}.\cite{zhao2023calibrating} and \cite{azar2024general} also directly optimize policies, and see \cite{ethayarajh2024kto} and \cite{xu2024contrastive} for related variants. Furthermore, \cite{azar2024general} proved the coincidence of optimal policies for RLHF and DPO.

\paragraph{Online Data Collection and On-Policy DPO.}
While the alignment frameworks are built upon a fixed dataset, they can be extended to an online setting in which the dataset is updated routinely. \cite{bai2022training} and \cite{touvron2023llama2} adopted an online RLHF setting, in which responses are generated from the current model, ranked by preference, and used to iteratively update the model through continual training on the newly collected data.  Responses can be ranked online using a separate LLM as in \cite{guo2024direct}, or the model itself being trained as in \cite{yuan2024selfreward}. \cite{xiong2024iterative} considered online data collection with iterative direct alignment. A more extensive review can be found in \cite{dong2024rlhf}.

A particularly promising direction in this line of work is on-policy implementation of DPO, which seeks to overcome the limitations of distributional mismatch inherent in fixed training datasets. Several studies \citep{yuan2024selfreward, chen2024self, guo2024direct, rosset2024direct, tajwar2024preference, pang2024iterative} demonstrate that sampling responses from intermediate models helps bridge the distributional gap and improves generalization. Empirical analyses by \citet{xu2024dpo} reveal that distributional mismatch between training data and the base model's original domain disproportionately impacts DPO compared to PPO. On-policy DPO actively samples the intermediate model generations, which serve as an adaptive distributional bridge to mitigate out-of-domain degradation. Despite the potential benefit of on-policy DPO, excessive reliance on the on-policy data is very likely to induce training instability that can lead to a significant drop in model performance \citep{lambert2024tulu3,deng2025less}. While trials have been made to balance the on-policy and off-policy data integration \citep{wang2025inco}, understanding when and how on-policy data can be helpful also remains to be further explored. 

Theoretical analyses of online DPO and the role of the data-generating distribution remain limited. The most closely related work is \citet{shi2024crucial}, which investigates the convergence rate of DPO under varying data-generating distributions. Our work differs in several key aspects. First, we provide theoretical results on the one-step gradient dynamics of the resulting policy, whereas \citet{shi2024crucial} treats gradient behavior as an assumption. Second, we analyze the influence of the data-generating distribution on DPO’s convergence behavior within a simplified linear alignment model, in contrast to \citet{shi2024crucial}, which make several assumptions about the full DPO framework to derive deeper convergence guarantees. For other, less directly related theoretical contributions, see \citet{feng2025pilaf} and \citet{huang2025can}.


\paragraph{Data Quality in LLM Alignment.} 
The importance of data quality in aligning large language models (LLMs) has been consistently demonstrated across various training paradigms. Early work by \cite{zhou2023lima} highlighted its decisive influence in fine-tuning settings, while recent advances in reasoning tasks \citep{muennighoff2025s1} further demonstrate the benefits of carefully curated alignment data. The quality of data can be assessed in several ways. \cite{wang2024arithmetic} adopted a multi-objective reward model to better gauge the value of data. \cite{liu2024skywork} evaluated the quality using LLMs as a reward model. \cite{cai2024internlm2} modified loss for finer appreciation of the quality.

There has been a surge of studies to obtain high quality data. \cite{fan202reformatted} enhanced the quality of instruction data by simple reformatting. \cite{cao2024instruction} and \cite{chen2024alpagasus} automated the process of premium data selection. \cite{lu2024instag} tagged supervised fine-tuning data to take diversity and complexity of the data into account.

In the context of DPO, empirical findings \citep{morimura2024filtered, wu2024beta, ivison2024unpacking} reveal two key insights: (1) DPO is more sensitive to data quality than traditional reinforcement learning methods such as PPO; and (2) the targeted selection of high-quality samples significantly enhances DPO performance. While some studies \citep{khaki2024rs, gou2024mixed} advocate for the use of larger preference gaps, others \citep{pattnaik2024enhancing, xiao2025finding} report that moderate gaps yield better outcomes. Despite these observations, a systematic understanding of how data quality influences DPO remains an open question in the literature.

\paragraph{Data Samplers in DPO.} Previous studies \citep{shi2024crucial,feng2025pilaf,xiong2024iterative,tajwar2024preference,guo2024direct} have explored how sampling schemes in online data collection can improve convergence rates or satisfy desirable first-order conditions. While their specific sampling strategies differ, they all implicitly enhance the quality of winning responses, which are closely aligned with the ``rewrite'' operation we emphasize. Among more closely related works, \cite{huang2024self} established convergence and sample complexity guarantees for the best-of-$K$ sampled DPO algorithm under theoretical assumptions, but did not address the online DPO setting, where the dataset evolves across iterations. \citet{deng2025less} showed that noise in preference data can degrade DPO performance, implicitly underscoring the importance of high-quality rewrites.

In contrast, our work explicitly focuses on the ``rewrite'' operation and its practical implications. We provide more qualitative and practitioner-relevant theoretical insights, highlighting that prior analyses often rely on strong theoretical assumptions, while in practice, challenges such as likelihood displacement and semantic correlation are prevalent. To bridge this gap, we introduce a simplified platform that disentangles some of the contradictions between theory and practice. Using this framework, we investigate the benefits of the rewrite operation and how it can be used to increase the probability of generating high-reward outputs.

\section{Experimental Details}\label{app:exp}
In Section \ref{section:experiment_setting1}, we introduce experimental settings and additional experimental results for online DPO on general LLMs. In Section \ref{section:experiment_setting2}, we state details of online DPO for the linear alignment model introduced in Section \ref{section:linear}. Section \ref{subsection:ref_model_impact} discusses about the experiment on the effect of reference model. Details of the experiments on the likelihood displacement are provided in Section \ref{appendix:likelihood_displacement}.

\subsection{Experiment Setting for LLM} \label{section:experiment_setting1}
\textbf{Base Model: } We employ Allen-AI's open-sourced \texttt{Llama-} \texttt{3.1-Tulu-3-8B-SFT} \citep{lambert2024tulu3} as the base model for DPO training. The model is exclusively supervised fine-tuned (SFT) on a mix of publicly available and transparent SFT data from Meta's official pre-trained model\citep{grattafiori2024llama3herdmodels}, making it possible to guarantee no data overlap during the SFT and DPO stages. 

\textbf{Dataset: } Based on the data of our base model, we select OpenBMB's \texttt{UltraFeedback} \citep{cui2024ultrafeedbackboostinglanguagemodels} as the dataset for DPO training and response improvement process. 

\textbf{Response Improvement Process: } We follow a structured response improving procedure for our online DPO setting. Given a prompt, we first generate 16 responses with the base model. Each response, including the original response, is then evaluated by \texttt{Skywork-Reward-Llama} \texttt{-3.1-8B-v0.2} \citep{liu2024skywork}, a reward model that assigns a score to a response. We update the preferred output by choosing the response with the highest score. The collected preference data are used for the current round of DPO, and the updated model serves as the base model in the next iteration.  

\textbf{Evaluation: } First, we assess the in- (ID) and out-of-distribution (OOD) performance of the trained model in figure \ref{fig:indist_distribution}. Given a prompt, we generate 8 responses from a model, assign scores with the reward model, and record the highest reward. For OOD data, we also report win rate in table \ref{tab:rewrite_winrate}, the proportion of prompts for which the trained model achieves a larger maximum reward than the base model. 


To further investigate whether our model exhibits \textit{reward hacking} behavior, we go beyond comparing win rates and evaluate the model on several standard benchmarks for large language models (LLMs). These benchmarks include \textbf{MMLU} \citep{hendrycks2021measuringmassivemultitasklanguage} , \textbf{TruthfulQA} \citep{lin2022truthfulqameasuringmodelsmimic} , \textbf{GSM8K} \citep{cobbe2021trainingverifierssolvemath} , and \textbf{IFEval} \citep{zhou2023instructionfollowingevaluationlargelanguage}. By comparing model performance we assess whether improvements in reward correspond to genuine gains in general capabilities or if they come at the cost of degraded downstream performance, signaling reward overoptimization.

\begin{table}[ht!]
\centering
\begin{tabular}{lcccc}
\toprule
\textbf{Model} & \textbf{MMLU} & \textbf{TruthfulQA} & \textbf{GSM8K} & \textbf{IFEval} \\
\midrule
$DPO_{base}$  & 0.6400 & 0.5894 & 0.8052 & 0.7616 \\
$DPO^{+}$   & 0.6292 & 0.5524 & 0.8082 & 0.7616 \\
$DPO^{++}$  & 0.6159 & 0.5710 & 0.7703 & 0.6861 \\
\bottomrule
\end{tabular}
\caption{Evaluation scores across tasks for different DPO-based models.}
\label{tab:other_DPO_bench}
\end{table}

\begin{table}[h]
\centering
\begin{tabular}{@{}lc@{}}
\toprule
\textbf{Comparison} & \textbf{Win Rate (\%)} \\
\midrule
$DPO^{+}$ vs. $DPO^{(1)}$ & 52.4 \\
$DPO^{++}$ vs. $DPO^{(2)}$ & 54.1 \\
\bottomrule
\end{tabular}
\captionof{table}{Pairwise win rates of models with and without the response improvement step.}
\label{tab:rewrite_winrate_app}
\end{table}

\subsubsection{Experimental Insights}

\textbf{No Sign of Reward Hacking.} Training exclusively with a specific reward model might lead to reward hacking \citep{skalse2022defining,pan2024feedback}, which occurs when a model exploits flaws in the reward signal to maximize reward in unintended ways, resulting in behavior misaligned with the true goals or values. The evaluation results in Table \ref{tab:other_DPO_bench} show no consistent evidence of reward hacking. The performance on general benchmarks remains largely stable. Particularly, GSM8K and MMLU scores show only minor fluctuations across rewrite strategies, and TruthfulQA performance does not degrade in a systematic way. Overall, the model validates that the improvements are not driven by reward-hacking behaviors.

\textbf{The Effect of Response Improvement Process.} To illustrate the effect of data improvement, we conduct a variant of Algorithm~\ref{alg:online_DPO} where the response improvement step (Step 6) is omitted. Specifically, between iterations we update only the reference policy while keeping $\mathcal{D}_t = \mathcal{D}$ fixed for all $t$. As shown in Table~\ref{tab:rewrite_winrate_app}, the resulting policy DPO$^{(1)}$ and DPO$^{(2)}$ (the superscript denotes the number of iterations for training the model) performs worse than the original Algorithm~\ref{alg:online_DPO} in terms of out-of-distribution win rate. This experiment further corroborates the importance of the data-generating distribution in the performance of DPO.

\subsubsection{Training Details for \texttt{Llama-3.1-Tulu-3-8B-SFT}\citep{lambert2024tulu3}}

For all DPO experiments, we adopt the standard DPO training pipeline using the Huggingface framework with the following hyperparameters:

\begin{itemize}
  \item \textbf{Optimizer:} Adam ($\beta_1 = 0.9$, $\beta_2 = 0.999$) with no weight decay
  \item \textbf{Learning Rate:} Cosine decay with linear warmup (warmup ratio = 0.1) to a peak of $5 \times 10^{-7}$
  \item \textbf{Batch Size:} A global batch size of 4 via gradient accumulation over 8 steps
  \item \textbf{Duration:} 3 epochs
  \item \textbf{DPO Beta:} 0.1
  \item \textbf{Sequence Length:} 2048
  \item \textbf{Precision:} \texttt{bfloat16}
  \item \textbf{Hardware}: $4 \times$ NVIDIA A100 80GB GPUs\quad 400GB Memory \quad 32 CPU cores 
    \item \textbf{Training time}: $\sim$76 hours
\end{itemize}

We use the default split provided by the dataset, which includes a predefined training set (in-distribution) and validation set (out-of-distribution). No additional manual partitioning was performed.

\subsection{Experiment Settings and Results on the Linear Alignment Model} \label{section:experiment_setting2}

\textbf{Base Model: } The model returns a linear transformation of its input $x$ with a Gaussian noise by
\[
    f(x) = w^\top x + \varepsilon, \quad \varepsilon \sim \mathcal{N}(0, \sigma^2).
\]
This formulation provides a controlled testbed to analyze DPO training in a simplified setting while allowing the model to include uncertainty of the predictions. 

\textbf{Dataset: } We construct a synthetic dataset. Specifically, we sample the input $x$ from the standard Gaussian distribution by $x \sim N_d(0,I)$ where $d$ is the input dimension. 

\textbf{Rewriting Process: } We follow a process similar to that of the LLM experiment. Given an input $x$, we first generate $K$ outputs using the base model. An output is scored by the $L^2$ distance to the true output of the reference model $f^*(x)=w^{*\top}x$ as
\[
r(x,y)=-||y-f^*(x)||_2^2.
\] 
Responses with the highest reward is selected as $y_1$ and we sample $y_2$ independently. Then we apply the BT model to label $y_w$ and $y_l$. Lastly, we perform DPO on the base model with the newly collected data, and the trained model acts as the base model in the next round.

\textbf{Evaluation: } Model performance is evaluated after every iteration. The performance can be evaluated in two ways. First, the performance can be measured by the $L^2$ distance between the trained model $f_{\theta}(\cdot)$ and $f^*(\cdot)$ on a fixed held-out evaluation set $\{x^{(j)}\}_{j=1}^m$ as
\begin{align*}
    \sum_{j=1}^m \left\| f_{\theta}(x^{(j)}) - f^*(x^{(j)}) \right\|_2^2.
\end{align*}
Also, we can evaluate the distance of the parameter $||w-w^*||^2$.
Smaller distance implies that the model is well aligned.

In addition, we conducted experiments with different initialization points to test the effect of starting conditions on alignment.
\begin{figure}[H]
    \centering
    \begin{subfigure}[t]{0.75\textwidth}
    \centering
    \includegraphics[width=\linewidth]{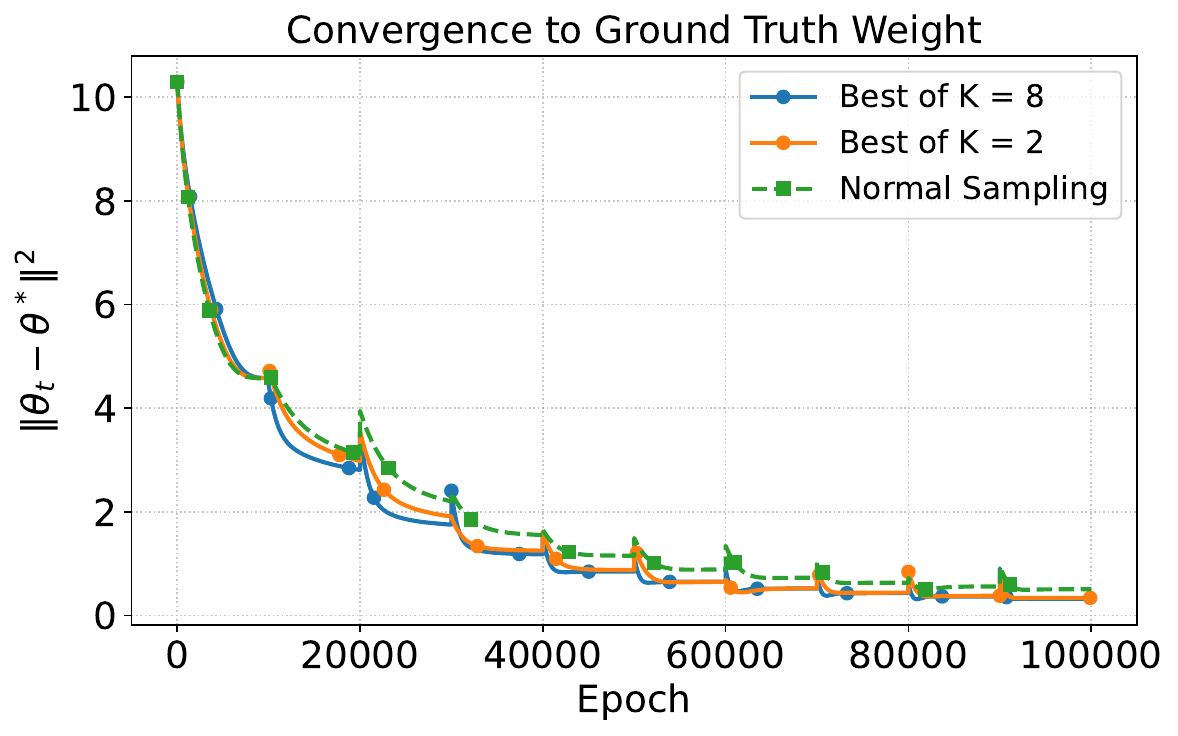}
        \label{fig:conv10}
    \end{subfigure}
    \caption{Convergence of Algorithm~\ref{alg:Data} to ground-truth weight in the linear alignment model from a different starting point.}
\end{figure}

\subsubsection{Experimental Insights}

\textbf{The Effect of Response Improvement Process.}
We compare several training regimes: standard sampling, best-of-$2$ sampling, and best-of-$8$ sampling. Models are identically initialized in both settings and use the same hyperparameters. As shown in Figure \ref{fig:gradient}, the model best-of-$8$ converges significantly faster towards the true parameter $w^*$. The distance between $w^*$ and the model parameter $w_t$ at step $t$ drops more rapidly, especially during the early training phase. The final error is also smaller compared to the baseline standard sampling. This corroborate the theoretical advantages illustrated in Proposition \ref{prop_fisher}.

\subsubsection{Training details for Synthetic Linear Model}
\label{Setting_Linear}

We simulate preference-based learning under a synthetic linear model to study DPO convergence dynamics with iterative model refinement. The following configuration is used:

\begin{itemize}
  \item \textbf{True Model:} Linear model $f^*(x) = w^{*\top} x+\epsilon$ with randomly sampled $w^* \in \mathbb{R}^{32}$
  \item \textbf{Data:} $2^{14}$ samples $x \sim \mathcal{N}(0, I_{32})$
  \item \textbf{Reference Model:} Initialized as a noisy perturbation of $w^*$.
  \item \textbf{Learned Model:} Initialized from the reference model with additional noise, trained using DPO loss
  \item \textbf{Reward Function:} $r(x,y) = -\|y - f^*(x)\|_2^2$
  \item \textbf{Preference Generation:} Two samples $(y_1, y_2)$ drawn from the model according to Algorithm \ref{alg:Data}, then labeled with BT model with $\tau=1.0$
  \item \textbf{Loss:} DPO loss with Gaussian log-likelihood
  \item \textbf{Optimizer:} Adam with learning rate $1 \times 10^{-4}$ 
  \item \textbf{Iterations:} 10,000 steps per phase; 10 total refinement phases
  \item \textbf{Metrics:} $\|w - w^*\|^2$ tracked across all iterations and exported for analysis
  \item \textbf{Hardware:} $1 \times$ NVIDIA A6000 GPU \quad 128GB Memory \quad 32 CPU cores 
  \item \textbf{Training time: }$\sim$ 30 minutes
\end{itemize}






\subsection{Experimental Details on the Effect of Reference Model} \label{subsection:ref_model_impact}

To complement the discussion in Section \ref{section:ref_sup}, we present empirical evidence demonstrating the influence of reference model on the capability of target policy. We compare two conditions: (1) a \emph{well-aligned reference model}, which produces predictions that are consistent with the true data distribution, and (2) a \emph{misaligned reference model}, which exhibits systematic deviations from the dataset.

\textbf{Reference Models.} We construct two reference policies by perturbing the ground-truth parameter $w^*$. 
\begin{align*}
    \text{Well-aligned reference: } \quad 
    w_{\text{ref}}^{\text{well}} &= w^* + \varepsilon_{\text{small}}, 
    &\varepsilon_{\text{small}} &\sim \mathcal{N}(0, \sigma_{\text{small}}^2 I_d), \\[1em]
    \text{Misaligned reference: } \quad 
    w_{\text{ref}}^{\text{mis}} &= w^* + \varepsilon_{\text{large}}, 
    &\varepsilon_{\text{large}} &\sim \mathcal{N}(0, \sigma_{\text{large}}^2 I_d),
\end{align*}
where $\sigma_{\text{large}}^2 \gg \sigma_{\text{small}}^2$. The misaligned reference lies much farther from the ground truth compared to the well-aligned reference. 

Both reference models induce the same form of conditional distribution over outputs:
\[
    y \mid x \;\sim\; \mathcal{N}\!\left((w_{\text{ref}}^{\text{well/mis}})^\top x, \, \sigma^2\right).
\]

The implied reference probability is
\[
    \pi_{\text{ref}}^{\text{well/mis}}(y \mid x) 
    = \frac{1}{\sqrt{2\pi\sigma^2}} 
      \exp\!\left(-\tfrac{1}{2\sigma^2}\big(y - (w_{\text{ref}}^{\text{well/mis}})^\top x \big)^2\right),
\]
offering a controllable way to adjust the probability density and verify our experiment.


\textbf{Learned Model Initialization.} The learned model is initialized from a fixed perturbation of ground-truth weights to ensure reproducibility across experimental runs. This design isolates the effect of the reference model. In particular, when using a \emph{well-aligned reference}, $\pi_{\text{ref}}$ assigns relatively high conditional probabilities to outputs with high rewards, leading to informative gradients and faster convergence. Formally, let the reward for an output $y$ be defined as $
r(y) = -\|y - y_{\text{gt}}\|_2^2$,
where $y_{\text{gt}}$ denotes the ground-truth output. A well-aligned reference model satisfies
\[
\pi_{\text{ref}}(y_{\text{chosen}} \mid x) > \pi_{\text{ref}}(y_{\text{reject}} \mid x)
\quad \text{whenever} \quad 
r(y_{\text{chosen}}) > r(y_{\text{reject}}),
\]
which implies that the model’s likelihoods are positively correlated with the reward function. This results in informative gradients under the DPO loss and accelerates convergence. 

In contrast, a \emph{misaligned reference} violates this correlation, i.e.,
\[
\pi_{\text{ref}}(y_{\text{chosen}} \mid x) < \pi_{\text{ref}}(y_{\text{reject}} \mid x)
\quad \text{for some pairs } (y_{\text{chosen}}, y_{\text{reject}}),
\]
causing the optimization to be less stable and hindering the learned model's ability to approximate the ground-truth distribution.

After DPO training under loss \eqref{DPO_form_f}, let $\pi_{\theta}^{\text{well}}(\cdot)$ and $\pi_{\theta}^{\text{mis}}(\cdot)$ denote the trained policies based on the reference models $\pi_{\text{ref}}^{\text{well}}(\cdot)$ and $\pi_{\text{ref}}^{\text{mis}}(\cdot)$, respectively. Let $w^{\text{well}}$ and $w^{\text{mis}}$ be the corresponding parameters of $\pi_{\theta}^{\text{well}}$ and $\pi_{\theta}^{\text{mis}}$. To evaluate the effect of using a misaligned reference model, we compare the following metrics:

\begin{itemize}
    \item \textbf{Average Conditional Probability of the Ground Truth.} We compute the average value of $\pi_{\theta}^{\text{well/mis}}(y_{\text{gt}})$, where $y_{\text{gt}}$ is the ground-truth response with the highest reward. This serves as a natural measure of the model’s likelihood of generating high-quality outputs, corresponding to Regions 1, 2, 6, and 7 in Figure~\ref{fig:support}.

    \item \textbf{Squared Distance to the Ground Truth.} We evaluate $\| w_{\theta}^{\text{well/mis}} - w^* \|^2$, which measures how closely the resulting model approximates the true model. This metric is strongly indicative of overall model quality.
\end{itemize}
We conduct two sets of experiments, using a well-aligned reference and a misaligned reference, respectively. Each configuration is run multiple times with different random seeds to account for variability, while keeping the target model initialization, dataset, and optimizer settings identical across runs. 

\textbf{Findings.} Our experiments highlight the impact of reference model quality on DPO training dynamics.

\begin{figure}[t]
\centering
\includegraphics[width=0.4\textwidth]{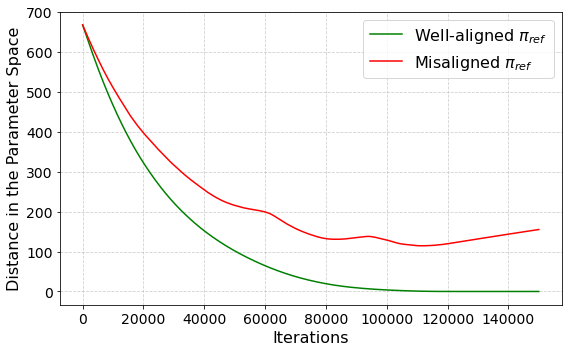}
\caption{Convergence of Model Parameters with Well-aligned and Misaligned Reference Models}
\label{fig:training_curve}
\end{figure}

\begin{figure}[h]
    \centering
    \begin{subfigure}[t]{0.46\textwidth}
        \centering
        \includegraphics[width=\linewidth]{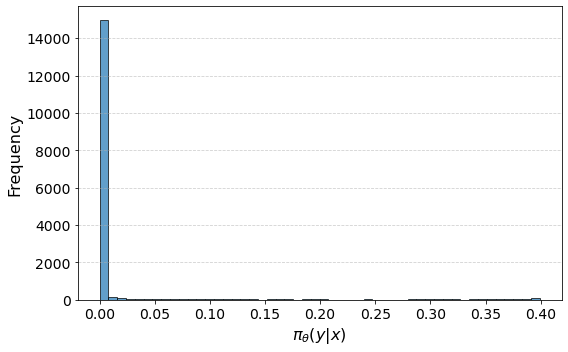}
        \caption{Collection of the predicted probabilities of the ground truth data $(x,y)$ before training the target policy with the well-aligned reference model. The performance of the target policy is poor before training.}
        \label{fig:start_well}
    \end{subfigure}
    \hfill
    \begin{subfigure}[t]{0.46\textwidth}
    \centering
    \includegraphics[width=\linewidth]{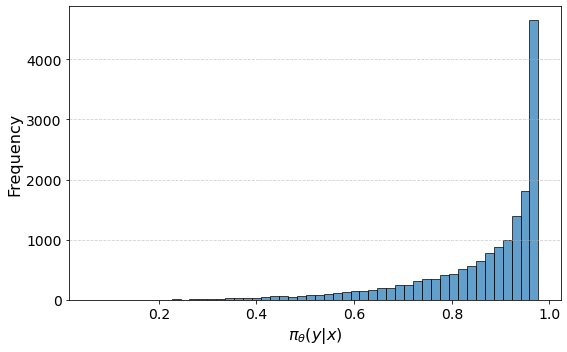}
        \caption{Predicted probabilities after training. The performance increased significantly after DPO training.}
        \label{fig:end_well}
    \end{subfigure}
    \caption{Change in Target Policy Before and After Training with Well-aligned Reference}
    \label{fig:ref_well}
\end{figure}

\begin{figure}[t]
    \centering
    \begin{subfigure}[t]{0.46\textwidth}
        \centering
        \includegraphics[width=\linewidth]{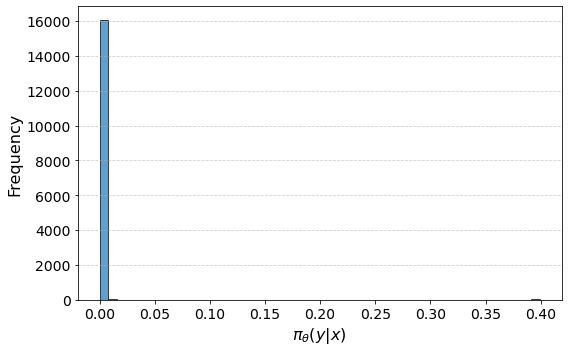}
        \caption{Collection of the predicted probabilities of the ground truth data $(x,y)$ before training the target policy with the misaligned reference model. The performance of the target policy is poor before training.}
        \label{fig:start_mis}
    \end{subfigure}
    \hfill
    \begin{subfigure}[t]{0.46\textwidth}
    \centering
    \includegraphics[width=\linewidth]{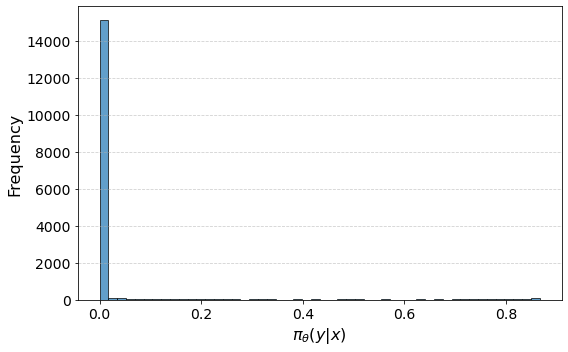}
        \caption{Predicted probabilities after training. The performance does not increase after training.}
        \label{fig:end_mis}
    \end{subfigure}
    \caption{Change in Target Policy Before and After Training with Misaligned Reference}
    \label{fig:ref_mis}
\end{figure}

\begin{itemize}
    \item \textbf{Well-aligned Reference:} When the reference model $\pi_{\text{ref}}$ is well-aligned, it assigns relatively high probability to ground-truth outputs, the learned policy exhibits fast and stable convergence, as shown in Figure~\ref{fig:training_curve}. In terms of the implied probability of high-reward outputs $y$, it increases substantially as shown in Figure~\ref {fig:ref_well}. Together, these trends demonstrate that a well-aligned reference provides informative gradients, guiding the model efficiently toward the target distribution.

    \item \textbf{Misaligned Reference:} When the reference model $\pi_{\text{ref}}$ assigns low probability to high-reward outputs, training becomes inefficient and may fail to converge to the optimal parameters. As shown in Figure~\ref{fig:training_curve}, the parameter distance $\|w - w^*\|^2$ decreases slowly and remains high throughout training, indicating limited progress toward the target weights. The DPO loss exhibits large fluctuations, and the learned model maintains a much lower likelihood for desirable outputs, as shown in Figure~\ref{fig:ref_mis}. These observations highlight that a poorly aligned reference provides weak or misleading gradient signals, which slow optimization and hinder convergence.
\end{itemize}



\subsubsection{Training details for Linear Model}
We use the same model and hyperparameter setting as \ref{Setting_Linear}

\begin{itemize}
  \item \textbf{Reference Model (Well-aligned):}  
  The reference model is initialized as a slightly perturbed version of the optimal weights $w^*$.  
  Specifically, the weights are defined as
  \[
  w = w^* + \delta, \quad \delta \sim \mathcal{N}(0, 0.05I),
  \]
  where $\delta$ represents a small Gaussian perturbation ensuring the model remains close to the optimal parameters.  
  This configuration represents a well-aligned reference model that starts near the true solution.  

\item \textbf{Reference Model (Misaligned):}  
The reference model is initialized to be far from the optimal weights $w^*$.  
A large random perturbation is applied to ensure substantial deviation from the target parameters.  
Specifically, the perturbation is generated as
\[
w = w^* + \delta,\quad  \delta \sim \mathcal{N}(0, 10I)
\]
Placing the model parameters at a large distance from $w^*$.  
This configuration represents a deliberately misaligned reference model, designed to start far from the optimal solution.

\end{itemize}

\subsection{Details for Gradient and Likelihood Displacement}
\label{appendix:likelihood_displacement}

This section provides experimental details for the empirical likelihood displacement results presented in Figure~\ref{fig:displacement}. The goal of this experiment is to investigate how DPO optimization shifts model likelihoods for different types of responses, under both isolated and aggregated update settings.

\paragraph{Gradient Update Modes.} For dataset $\mathcal{D} = \left\{\left(x^{(i)},y_w^{(i)},y_l^{(i)}\right)\right\}_{i=1}^n$, with any parameter $\theta_t$, the batch gradient descent takes the form
\begin{align*}
    \theta_{t+1} &= \theta_t - \eta \nabla_{\theta}\hat{\mathcal{L}}(\theta_t,\mathcal{D})\\
    &=\theta_t - \eta \sum_{i=1}^n\nabla_{\theta}\ell(\theta_t,(x^{(i)},y_w^{(i)},y_l^{(i)})),
\end{align*}
where $\ell(\theta,(x,y_w,y_l)) = -\log \sigma\left(f_\theta(x, y_w) - f_\theta(x, y_l)\right)$. We consider two update schemes:

\begin{itemize}
\item \textbf{Independent Gradient:} To isolate the influence of individual preference samples, that is, the gradient effect of $(x^{(i)},y_w^{(i)},y_l^{(i)})$ might interfere with the effect of $(x^{(j)},y_w^{(j)},y_l^{(j)})$ for $i\neq j$, we first evaluate the gradient descent effect of each sample and see how the policy will change. For all $(x^{(i)},y_w^{(i)},y_l^{(i)}) \in \mathcal{D}$, we independently compute $\theta_{t+1}^{(i)}$ via
\begin{align*}
    \theta_{t+1}^{(i)} = \theta_t - \eta \nabla_{\theta}\ell(\theta_t,(x^{(i)},y_w^{(i)},y_l^{(i)})),
\end{align*}
and evaluate the change $\pi_{\theta_{t+1}^{(i)}}(\cdot|x)$ compared to $\pi_{\theta_{t}}(\cdot|x)$. Specifically, we compute $$\Delta_{\theta_{t+1}^{(i)}}(x,y) = f_{\theta_{t+1}^{(i)}}(x, y) - f_{\theta_t}(x, y) \propto \log\left(\pi_{\theta_{t+1}^{(i)}}(y|x)/\pi_{\theta_{t}}(y|x)\right),$$ and record this value as the sample’s immediate effect on the model. For Figure \ref{fig:displacement_linear}, we plot the distributional property of $\left\{\Delta_{\theta_{t+1}^{(i)}}(x^{(i)},y)\right\}_{i=1}^n$ for various type of $y$s that we will discuss later.
    
\item \textbf{Cumulative Effect:} To evaluate the cumulative effect, instead of evaluating $\theta_{t+1}^{(i)}$ for every $i$, we directly update $\theta_{t+1}$ via batch gradient descent
\begin{align*}
    \theta_{t+1} =\theta_t - \eta \sum_{i=1}^n\nabla_{\theta}\ell(\theta_t,(x^{(i)},y_w^{(i)},y_l^{(i)})).
\end{align*}
After the batch update, we compute the difference between each responses $\Delta_{\theta_{t+1}}(x,y) = f_{\theta_{t+1}}(x, y) - f_{\theta_{t}}(x, y).$ This setup reflects the cumulative effect of a batch of samples jointly influencing the model through gradient averaging. We plot the distributional property of $\left\{\Delta_{\theta_{t+1}}(x^{(i)},y)\right\}_{i=1}^n$ for various type of $y$s that we will discuss later.

\end{itemize}

\paragraph{Evaluating Likelihood Change.}  
To study how model likelihoods change under both update schemes, we expand the dataset to incorporate more response. For each prompt $x\in\{x^{(i)}\}_{i=1}^{64}$, we curate four distinct types of responses:

\begin{itemize}
    \item \textbf{Chosen Response ($y_{\text{w}}|x$):} The preferred response as determined by ChatGPT-4O in response to $x$.
    
    \item \textbf{Rejected Response ($y_{\text{l}}|x$):} A dispreferred response also generated by ChatGPT-4O, which is explicitly contrasted with $y_{\text{w}}$ in the preference pair.
    
    \item \textbf{Independent Response ($y_{\text{ind}}|x$):} A third response independently generated by ChatGPT-4O and then curated by human, ensuring semantic and stylistic diversity from both $y_{\text{w}}$ and $y_{\text{l}}$.
    
    \item \textbf{Correlated Response ($y_{\text{cor}}|x$):} A manually constructed variant of the preferred response $y_{\text{w}}$, obtained by making token-level edits (e.g., synonym substitution, paraphrasing, punctuation changes) while preserving the overall semantics.
\end{itemize}




\paragraph{Experimental Insights.}  
Our experiments reveal that while DPO updates perform as expected in isolated cases, semantic correlations among responses introduce notable complexity when training on larger datasets. In single-sample \textbf{Independent case}, one step of gradient descent increases the likelihood of the preferred response $y_w$ and decrease the less preferred response $y_l$ with negligible effect on unrelated responses $y_{ind}$. However, semantically similar responses $y_{cor}$ also exhibit increased likelihood, suggesting that the model's internal representation space propagates preference signals beyond the directly compared pair. 

When scaling to datasets with \textbf{cumulative preference tuples}, we observe that gradients from one tuple influence semantically related responses in other tuples, causing gradient interference. We call this 
\textbf{semantic spillover effects}, and can be a potential reason leading to \textbf{likelihood displacement}, when the probability of preferred response $y_w$  decreases after training. Although each tuple's gradient aligns with the theoretical expectation (Theorem \ref{thm:empirical_GD}), their aggregation can diverge due to entangled representation structure. This observation underscores the limitations of treating preference tuples as independent in natural language tracks and motivates further refinement of training objectives that account for semantic entanglement.

\section{Technical Lemmas and Observations}\label{app_A}
In this section, we list technical lemmas needed to prove results provided in the main text and appendix. All the statements will be proved in Appendix \ref{section:proof}.

\subsection{Lemmas for Section \ref{section:prelim}}

The generation of two responses $Y_1$ and $Y_2$ and the process of labeling the samples as preferred $Y_w$ or dispreferred $Y_l$ are connected as below.
\begin{lemma}\label{lemma_density}
We have 
\begin{equation}
    \begin{aligned}
        p_{Y_w,Y_l|x}(y,y'|x) &= (p_{Y_1,Y_2|x}(y, y'|x) + p_{Y_1,Y_2|x}(y', y|x)) \sigma(r^*(x,y) - r^*(x,y')),
    \end{aligned}
\end{equation}
where the first term is the probability (density) of generating an unordered pair (y, y'), assuming there is no identical pair; the second term is the probability that $y$ is preferred to $y'$.
\end{lemma}

The DPO loss $\mathcal{L}$ is built on the preferred and dispreferred samples $(y_w,y_l)$. The following lemma symmetrizes the loss by expressing it in terms of unordered pair $(x,y,y^\prime) \sim \mathcal{D}_u$.
\begin{lemma}\label{lem_DPO_loss}
    Under the DPO loss, the gradient has an alternative form
    \begin{align*}
        \nabla_{\theta}\mathcal{L}(\theta) &= -\mathbb{E}_{(x,y,y') \sim \mathcal{D}_u}\left[ \{\sigma(r^*(x, y) - r^*(x,y')) - \sigma(f_{\theta}(x,y) - f_{\theta}(x, y')) \}\right.\\&\hspace{40mm}\left.\cdot(\nabla_{\theta}f_{\theta}(x, y) - \nabla_{\theta}f_{\theta}(x,y'))\right].
    \end{align*}
\end{lemma}

\subsection{Lemmas for Section \ref{section:linear}}

In order to analyze the best-of-$K$ sampling introduced in Section \ref{section:convergence}, the distribution of the best data chosen among $K$ samples should be specified. The following lemma gives the precise distribution of the noise corresponding to the best sample. Note that the standard sampling in Algorithm \ref{alg:Data} is a special case of the best-of-$K$ sampling with $K=1$, hence all the analyses on general $K$ apply to the standard sampling as well.

\begin{lemma} \label{lemma:resample_pdf}
    For a fixed input $x \in \mathbb{R}^d$, suppose responses $y^{(i)}$ $(i=1, \cdots, K)$ are generated by $y^{(i)} = w^\top x + \sigma \epsilon^{(i)}$ for some $w \in \mathbb{R}^d$ where $\epsilon^{(i)} \overset{iid}{\sim} \mathcal{N}(0,1)$ are Gaussian noises. Choose $m = \argmin_{i} |y^{(i)}-w^{*\top}x|$ to be the sample that is closest to the target model $y=w^{*\top}x$ and denote $\epsilon_0 := \epsilon^{(m)}$. Then the distribution of $\epsilon_0$ is 
    \begin{align*}
        p_{\epsilon_0}(u) = K \phi(u) \left\{1-F_{|\delta(x)+Z|}(|\delta(x)+u|) \right\}^{K-1}
    \end{align*}
    where $\phi$ is the PDF of the standard normal distribution, $\delta(x) = \frac{(w-w^*)^\top x}{\sigma}$, and $F_{|\delta(x)+Z|}$ is the CDF of $|\delta(x)+Z|$ with $Z \sim \mathcal{N}(0,1)$.
\end{lemma}

\textbf{Notation Change.} As introduced in Section \ref{section:linear}, a Gaussian policy $\pi_\theta$ is parametrized by $\theta = (w,\sigma)$. With this choice of model, the gradient and the Hessian of the DPO loss can be concretely expressed as below. In the following lemmas and their corresponding proofs, let us denote the preference data by $(x,y_+,y_-)$ instead of $(x,y_w,y_l)$, because in the linear alignment model $w$ represents both the preferred response and the weight of the Gaussian model, which might cause confusion. Similarly, we denote the logistic function by $\tilde{\sigma}$ to distinguish it from the standard deviation $\sigma$ of the Gaussian model.

\begin{lemma}\label{lemma:DPO_loss_grad}
    Let $(x,y_+,y_-)$ be the preference data. Denote 
\begin{align*}
    \ell(\theta; x,y_+, y_-) = -\log{\tilde{\sigma}(f_\theta(x,y_+)-f_\theta(x,y_-))}
\end{align*}
where $f_\theta(x,y) = \beta \log{\frac{\pi_\theta(y|x)}{\pi_{ref}(y|x)}}$. For the Gaussian models $\pi_\theta(\cdot|x) = \mathcal{N}(w^\top x, \sigma^2)$ and $\pi_{ref}(\cdot|x) = \mathcal{N}(w_{ref}^\top x, \sigma_{ref}^2)$, the gradient and the Hessian of $\ell$ are 
\begin{align}
    \nabla_w \ell  
    &= -\beta\frac{\sigma_{ref}}{\sigma^2} \left\{1-\tilde{\sigma}(f_\theta(x,y_+)-f_\theta(x,y_-))  \right\} (\epsilon_+ - \epsilon_-) x , \label{eq:grad} \\
    \nabla_w^2 \ell 
    &= \beta^2 \frac{\sigma_{ref}^2}{\sigma^4 }( \tilde{\sigma} \cdot (1-\tilde{\sigma}) ) (f_\theta(x,y_+)-f_\theta(x,y_-)) \left( \epsilon_+ - \epsilon_- \right)^2 x x^\top, \label{eq:grad3}
\end{align}
where $\epsilon_+ = \frac{y_+-w_{ref}^\top x}{\sigma_{ref}}$ and $\epsilon_- = \frac{y_- - w_{ref}^\top x}{\sigma_{ref}}$.
\end{lemma}

\subsection{Observations from Section \ref{section:linear}}

As noted in Section \ref{section:convergence}, understanding $\gamma(K,\delta(x)) = \mathbb{E}[|\epsilon_1-\epsilon_2|]$ and $\eta(K,\delta(x)) = \mathbb{E}[\epsilon_1^2]$ is crucial for analyzing the first and the second order properties of the training dynamics of DPO. The two functions $\gamma$ and $\eta$ depend on the bias $\delta(x)$. In the following, we provide approximates of $\gamma$ and $\eta$ on large and small $|\delta(x)|$ regimes, respectively.

\begin{observation}\label{observation:approximation}
    Let $\gamma(K,\delta(x))$ and $\eta(K,\delta(x))$ be defined as in Lemma \ref{lem:analytical}. For large enough $|\delta(x)|$, the order of the two functions are
    \begin{align*}
        \gamma(K,\delta(x)) = O(|\delta(x)|) \quad \text{and} \quad \eta(K,\delta(x)) = O(\delta(x)^2).
    \end{align*}
\end{observation}

\begin{observation}\label{observation:approximation2}
    The functions $\gamma(K,\delta(x))$ and $\eta(K,\delta(x))$ in Lemma \ref{lem:analytical} are bounded as follows for small $|\delta(x)|$:
    \begin{align*}
        \lim_{\delta(x) \rightarrow 0} \gamma(K,\delta(x)) \geq \sqrt{\frac{2}{\pi}} 
        \quad \text{and} \quad 
        \lim_{\delta(x) \rightarrow 0} \eta(K,\delta(x)) \leq \frac{1}{2}.
    \end{align*}
\end{observation}

\section{Proofs} \label{section:proof}

\subsection{Proofs Related to Section \ref{section:RLHF_sol}} \label{subsection:proof_supp}

In this section, Proposition \ref{prop_optimal_loss}, Lemma \ref{lem:redefine}, and Observation \ref{prop:DPO_supp} are proved. We also introduce and prove Observation \ref{obs:RLHF_supp}, a finding for RLHF trained models, which is analogous to Observation \ref{prop:DPO_supp} for DPO trained models.

\subsubsection{Proof of Proposition \ref{prop_optimal_loss}}

\begin{repproposition}{prop_optimal_loss}
(restated) The function $f^*(x,y) = r^*(x,y) + c(x)$, where $c(x)$ is a function of $x$ only, is a global optimal solution to \eqref{DPO_form_f}. Consequently, the policy $\pi^*(y|x)$ defined as 
\begin{align*}
    \pi^{*}(y|x) = \frac{1}{Z(x)}\pi_{\text{ref}}(y|x)\exp\left(\frac{1}{\beta}r^*(x,y)\right),
\end{align*}
where $Z(x) = \int \pi_{\text{ref}}(y|x)\exp\left(\frac{1}{\beta}r^*(x,y)\right)dy$, is optimal solution for both the RLHF formulation \eqref{RLHF_original} and the DPO formulation \eqref{DPO_BT_definition}.
\end{repproposition}

\begin{proof}
The DPO loss is defined as
\[
    \mathcal{L}(\theta) = -\mathbb{E}_{(x, y_w, y_l) \sim \mathcal{D}} \left[ \log \sigma \left( f_\theta(x, y_w) - f_\theta(x, y_l) \right) \right].
\]

Lemma \ref{lem_DPO_loss} gives the following form of the gradient of the loss:
\begin{align}\label{eq:loss_grad}
\begin{split}
    \nabla_\theta \mathcal{L}(\theta) &= -\mathbb{E}_{(x, y, y') \sim \mathcal{D}_u} \big[
    \{\sigma\left( r^*(x, y) - r^*(x, y') \right) - \sigma\left( f_\theta(x, y) - f_\theta(x, y') \right) \} \\
    & \quad \times \left( \nabla_\theta f_\theta(x, y) - \nabla_\theta f_\theta(x, y') \right)\big].
\end{split}
\end{align}
For \( f^*(x, y) = r^*(x, y) + c(x) \), where \( c(x) \) is a function of \( x \) only,
\begin{align} \label{eq:same_sigmoid}
    \sigma\left( f^*(x, y) - f^*(x, y') \right) = \sigma\left( r^*(x, y) - r^*(x, y') \right)
\end{align}
holds. Therefore, evaluating $\nabla_\theta \mathcal{L}$ at $f^*$ by substituting \eqref{eq:same_sigmoid} into \eqref{eq:loss_grad} gives 
\[
\nabla_\theta \mathcal{L}(f^*)  = 0,
\]
which implies that $f^*$ is a stationary point of the DPO loss.

Moreover, the difference $z = f(x,y)-f(x,y')$ is linear in $f$. Since the function $\psi(\cdot)=-log(\sigma(\cdot))$ is convex, and $z$ is linear in $f$, the DPO loss is convex with respect to $f_\theta$.
As a result, the stationary point $f^*$ achieves the global minimum.


To prove that the optimal policy \eqref{RLHF_sol} of \eqref{RLHF_original} is a global minimizer of \eqref{DPO_BT_definition}, rearrange \eqref{RLHF_sol} to obtain
\begin{align*}
    \beta \log{\frac{\pi^*(y|x)}{\pi_{ref}(y|x)}} = r^*(x,y) -\beta \log{Z(x)}.
\end{align*}
Since the relative logit corresponding to $\pi^*$ is
\begin{align*}
    f_{\pi^*}(x,y) = \beta \log{\frac{\pi^*(y|x)}{\pi_{ref}(y|x)}},
\end{align*}
it follows that 
\begin{align*}
    f_{\pi^*}(x,y) = r^*(x,y) -\beta \log{Z(x)},
\end{align*}
which is of the form $r^*(x,y) + c(x)$ where $c$ is a function of $x$ only. Therefore, $\pi^*$ is an optimal policy for \eqref{DPO_BT_definition} by the previous result.

\end{proof}

\subsubsection{Proof of Lemma \ref{lem:redefine}}

\begin{replemma}{lem:redefine}
(restated) Let $\pi_\theta$ be a minimizer for \eqref{DPO_BT_definition}. Then $\pi_\theta^\prime$ defined as below is also a minimizer for \eqref{DPO_BT_definition}:
\begin{align*}
    \pi_\theta^\prime(y|x) = \begin{cases} \pi_\theta(y|x) \phi(x) & \text{if } y \in supp(\mathcal{D}_{Y|x}) \\ \tilde{\pi}_\theta (y|x) & \text{otherwise} \end{cases},
\end{align*}
where $\phi$ is a function of $x$ only that satisfies $\phi(x) \in \left(0, 1/\int_{supp(\mathcal{D}_{Y|x})} \pi_\theta(y|x) dy\right)$, and $\tilde{\pi}_\theta(\cdot|\cdot)$ is any policy such that $\int_{supp(\mathcal{D}_{Y|x})^c} \tilde{\pi}_\theta (y|x) dy = 1 - \phi(x)  \int_{supp(\mathcal{D}_{Y|x})} \pi_{\theta} (y|x) dy$.
\end{replemma}
\begin{proof}
The conditions on $\phi$ and $\tilde{\pi_\theta}$ give
\begin{align*}
    \int \pi_\theta^\prime (y|x) dy 
    &= \int_{supp(\mathcal{D}_{Y|x})} \pi_\theta^\prime (y|x)dy  +  \int_{supp(\mathcal{D}_{Y|x})^c}\pi_\theta^\prime (y|x)dy \\
    &= \int_{supp(\mathcal{D}_{Y|x})} \pi_\theta (y|x) \phi(x)dy  +  \int_{supp(\mathcal{D}_{Y|x})^c}\tilde{\pi_\theta} (y|x)dy \\
    &= 1,
\end{align*}
which proves that $\pi_\theta^\prime$ is a policy.
Now, as in the proof of Proposition \ref{prop_optimal_loss},
\begin{align*}
    \sigma\left(\beta\left(\log{\frac{\pi_\theta^\prime(y_w|x)}{\pi_{\text{ref}}(y_w|x)}}-\log{\frac{\pi_\theta^\prime(y_l|x)}{\pi_{\text{ref}}(y_l|x)}}\right)\right)
    &= \sigma\left(\beta\left(\log{\frac{\pi_\theta(y_w|x)}{\pi_{\text{ref}}(y_w|x)}}+\log{\phi(x)}-\log{\frac{\pi_\theta(y_l|x)}{\pi_{\text{ref}}(y_l|x)}}-\log{\phi(x)}\right)\right) \\
    &= \sigma\left(\beta\left(\log{\frac{\pi_\theta(y_w|x)}{\pi_{\text{ref}}(y_w|x)}}-\log{\frac{\pi_\theta(y_l|x)}{\pi_{\text{ref}}(y_l|x)}}\right)\right) 
\end{align*}
for $y_w, y_l \in supp(\mathcal{D}_{Y|x})$. Hence the DPO losses of $\pi_\theta$ and $\pi_\theta^\prime$ are the same, which implies that $\pi_\theta^\prime$ is also an optimal policy for the DPO loss.
\end{proof}

\subsubsection{Proof of Observation \ref{obs:RLHF_supp}}

\begin{observation} \label{obs:RLHF_supp}
A maximizer $\pi_{\theta^*}$ for \eqref{RLHF_original}  is positive only on the region where $\pi_{\text{ref}}(y|x)$ is positive.
\end{observation}

\begin{proof}
To begin with, the maximum value of the RLHF objective \eqref{RLHF_original} is not $-\infty$ as plugging in $\pi_{\text{ref}}$ to $\pi_\theta$ gives
\begin{align*}
    \max_{\pi_\theta} \mathbb{E}_{x \sim \mathcal{D}_X, y \sim \pi_\theta(\cdot|x)} \left[r^*(x,y) - \beta KL(\pi_\theta(\cdot|x) || \pi_{\text{ref}}(\cdot|x))\right] 
    &\geq \mathbb{E}_{x \sim \mathcal{D}_X, y \sim \pi_{\text{ref}}(\cdot|x)} \left[r^*(x,y) - \beta KL(\pi_{\text{ref}}(\cdot|x) || \pi_{\text{ref}}(\cdot|x))\right] \\
    &= \mathbb{E}_{x \sim \mathcal{D}_X, y \sim \pi_{\text{ref}}(\cdot|x)} \left[r^*(x,y)\right].
\end{align*}
Hence we may consider only the policies that satisfy $KL(\pi_\theta (\cdot |x) || \pi_{\text{ref}} (\cdot|x)) \neq \infty$ as optimal solution. 

A well known property of KL divergence is that for two distributions $p_1$ and $p_2$, the KL divergence $KL(p_1||p_2)$ is $\infty$ only if there exists a point $z$ such that $p_1(z)\neq 0$ and $p_2(z) =0$. Therefore, a policy $\pi_\theta$ should be $0$ at $y$ if $\pi_{\text{ref}}(y|x) = 0$, 
or equivalently, $supp(\pi_\theta(\cdot | x)) \subset supp(\pi_{\text{ref}}(\cdot | x))$ for all $x \in supp(\mathcal{D}_X)$.
\end{proof}

\subsubsection{Proof of Observation \ref{prop:DPO_supp}}

\begin{repobservation}{prop:DPO_supp}
(restated) Let $\pi_\theta$ be a minimizer for \eqref{DPO_BT_definition}. Then on $supp(\mathcal{D}_{Y|x})$, we have \\
\hspace*{5pt} i) $\pi_\theta(y|x)>0$ if $\pi_{\text{ref}}(y|x)>0$ ; \quad\quad ii) $\pi_\theta(y|x)=0$ if $\pi_{\text{ref}}(y|x)=0.$
\end{repobservation} 

\begin{proof}
Similar to the proof of Observation \ref{obs:RLHF_supp}, the minimum value of the DPO loss \eqref{DPO_form_f} is not $\infty$ as plugging in $\pi_{\text{ref}}$ to $\pi_\theta$ gives
\begin{align*}
    \min_{\theta} \mathcal{L} 
    &= \min_{\theta} -\mathbb{E}_{(x,y_w,y_l) \sim \mathcal{D}}\left[\log{\sigma\left(\beta \left( \log{\frac{\pi_\theta(x,y_w)}{\pi_{\text{ref}}(x,y_w)}} - \log{\frac{\pi_\theta(x,y_l)}{\pi_{\text{ref}}(x,y_l)}}\right)\right)}\right] \\
    &\leq -\mathbb{E}_{(x,y_w,y_l) \sim \mathcal{D}}\left[\log{\sigma \left(\beta\left( \log{\frac{\pi_{\text{ref}}(x,y_w)}{\pi_{\text{ref}}(x,y_w)}} - \log{\frac{\pi_{\text{ref}}(x,y_l)}{\pi_{\text{ref}}(x,y_l)}}\right)\right)}\right] \\
    &=0.
\end{align*}
Hence we may consider only the policies that satisfy $\log{\frac{\pi_{\theta}(y_w|x)}{\pi_{\text{ref}}(y_w | x)}} \neq -\infty$, which happens if $\pi_\theta(y_w|x) = 0$ and $\pi_{\text{ref}}(y_w|x) \neq 0$. Therefore, a policy $\pi_\theta$ should be nonzero at a preferred response $y_w$ whenever $\pi_{\text{ref}}(y_w|x) \neq 0$. Under the BT model, any response $y$ in $\mathcal{D}$ has a positive probability of being a preferred response over any other response $\tilde{y}$. Hence $supp(\pi_{\text{ref}}(\cdot | x)) \cap supp(\mathcal{D}_{Y}) \subset supp(\pi_\theta (\cdot|x)) \cap supp(\mathcal{D}_{Y})$  for all $x \in supp(\mathcal{D}_X)$.

By the same reasoning as above, we may consider only the policies that satisfy $\log{\frac{\pi_{\theta}(y_l|x)}{\pi_{\text{ref}}(y_l | x)}} \neq \infty$, which happens if $\pi_\theta(y_l|x) \neq 0$ and $\pi_{\text{ref}}(y_l|x)= 0$. Therefore, $\pi_\theta$ should be $0$ at a dispreferred response $y_l$ if $\pi_{\text{ref}}(y_l|x)=0$. As any response can be a dispreferred response, it follows that $supp(\pi_{\text{ref}}(\cdot | x)) \cap supp(\mathcal{D}_{Y}) \supset supp(\pi_\theta (\cdot|x)) \cap supp(\mathcal{D}_{Y})$.
\end{proof}

\subsection{Proofs Related to Section \ref{section:gradient}}

Theorem \ref{thm:gradiennt} is proved in this section. In addition, Theorem \ref{thm:empirical_GD}, and empirical version of Theorem \ref{thm:gradiennt}, is also stated and proved.

\subsubsection{Proof of Theorem \ref{thm:gradiennt}} \label{pf:thm:gradiennt}

\begin{reptheorem}{thm:gradiennt}
(restated) Let $\pi_{\theta_{t+1}}$ be the policy after one step of gradient descent on $\mathcal{L}(\theta_t,\mathcal{D})$ defined in \eqref{DPO_BT_definition} at step $t$. Then the updated policy is compared with the previous policy as 
\begin{enumerate}[i)]
\item $\pi_{\theta_{t+1}}(y|x) >  \pi_{\theta_t}(y|x) \,\,\,\text{if } ~ \mathbb{P}_w(y|x) > \mathbb{P}_{w,\theta_t}(y|x)$
\item $ \pi_{\theta_{t+1}}(y|x) =  \pi_{\theta_t}(y|x) \,\,\,\text{if } ~ \mathbb{P}_w(y|x) = \mathbb{P}_{w,\theta_t}(y|x)$ or $y \notin \text{supp}(\mathcal{D}_{Y|x})$
\item $\pi_{\theta_{t+1}}(y|x) <  \pi_{\theta_t}(y|x) \,\,\,\text{if } ~ \mathbb{P}_w(y|x) < \mathbb{P}_{w,\theta_t}(y|x)$.
\end{enumerate}
Moreover, the change in the log-probability $f_{\theta_{t+1}}(y) - f_{\theta_t}(y)$ is proportional to the probability gap $\mathbb{P}_w(y|x) - \mathbb{P}_{w,\theta_t}(y|x)$ up to $O(\alpha^2)$ by 
\begin{align*}
    f_{\theta_{t+1}} (x,y) - f_{\theta_t}(x,y) &= 2\alpha  \left(\mathbb{P}_w(y|x) - \mathbb{P}_{w,\theta_t}(y|x)\right)p_{X,Y_1}(x,y) \left.\frac{\partial f_\theta}{\partial \theta}\right\vert_{\theta=\theta_t}(x,y)^2 + O(\alpha^2).
\end{align*}
\end{reptheorem}

\begin{proof}
For brevity, let us denote the relative logit of $\pi_\theta$ and $\pi_{\text{ref}}$ by
$f_{\theta}(x,y) = \beta \log\frac{\pi_{\theta}(y|x)}{\pi_{\text{ref}}(y|x)}$. Then
\begin{align*}
    \mathcal{L}(\theta,\mathcal{D}) = \mathbb{E}_{(x,y_w,y_l)\sim\mathcal{D}}[-\log(\sigma(f_{\theta}(x,y_w) - f_{\theta}(x,y_l)))] .
\end{align*}
With the gradient update  
\begin{align}\label{eqn_gradient}
    \theta_{t+1} = \theta_t - \alpha \cdot \nabla_\theta \mathcal{L} (\theta_t,\mathcal{D}),
\end{align}
the two policies $\pi_{\theta_{t+1}}(y|x)$ and $\pi_{\theta_{t}}(y|x)$ have the following relationship 
\begin{align}\label{eq:policy_GD}
    f_{\theta_{t+1}}(x,y) = f_{\theta_t}(x,y) - \alpha g_{\theta_t}(x,y) + O(\alpha^2)
\end{align} 
via second-order approximation where $g_\theta(x,y)$ is proportional to the functional derivative of $\mathcal{L}$ with respect to $f_\theta$ by $g_\theta (x,y) = \frac{\delta \mathcal{L}}{\delta f_\theta} (x,y) \frac{\partial f_\theta}{\partial \theta}(x,y)^2$. Note that $\alpha$ is usually taken around $10^{-5}$, so $O(\alpha^2)$ is negligible.
In order to compute $\frac{\delta \mathcal{L}}{\delta f}$, we explicitly express the dependency of $\mathcal{L}$ to $f_\theta$ as $\mathcal{L}[f_\theta]$. Since  
\begin{align*}
    \mathcal{L}[f + \epsilon \tilde{f}] 
    = \mathbb{E}_{(x, y_w, y_l) \sim \mathcal{D} } \left[-\log{\sigma\left(f(x,y_w) + \epsilon \tilde{f}(x,y_w) - f(x,y_l) - \epsilon\tilde{f}(x,y_l)  \right) }\right]
\end{align*} 
for a test function $\tilde{f}$, the functional differential is
\begin{align}\label{ftldifferential}
    \delta \mathcal{L} [f,\tilde{f}] 
    &= \left. \frac{\partial}{\partial \epsilon}\mathcal{L}[f + \epsilon \tilde{f}] \right\vert_{\epsilon=0} \nonumber \\ 
    &= - \mathbb{E}_\mathcal{D} \left[ \left\{1- \sigma\left(f(x,y_w) - f(x,y_l) \right)\right\} (\tilde{f}(x,y_w) - \tilde{f}(x,y_l)) \right].
\end{align}
Let us define $q(y_1,y_2|x) = \frac{1}{2}\left(p_{Y_1,Y_2|X}(y_1,y_2|x)+p_{Y_1,Y_2|X}(y_2,y_1|x)\right)$. Then \eqref{ftldifferential} extends to 
\begin{align*}
    \delta \mathcal{L} [f,\tilde{f}] 
    &= -\iiint \frac{e^{-(f_\theta(x,y_1) - f_\theta(x,y_2))}}{1+e^{-(f_\theta(x,y_1) - f_\theta(x,y_2))}} (\tilde{f}(x,y_1) - \tilde{f}(x,y_2)) p_X(x) \left(p_{Y_1, Y_2|X}(y_1, y_2|x)+p_{Y_1, Y_2|X}(y_2, y_1|x)\right) \\
    &\quad \times \mathbb{P}(y_1 \succ y_2 | x) dy_2 dy_1 dx \\
    &= \iint - \int\frac{e^{-(f_\theta(x,y_1) - f_\theta(x,y_2))}}{1+e^{-(f_\theta(x,y_1) - f_\theta(x,y_2))}} p_X(x) 2q(y_1, y_2|x) \mathbb{P}(y_1 \succ y_2 | x) dy_2 \tilde{f}(x,y_1)  dy_1 dx \\
    &\quad + \iint  \int\frac{e^{-(f_\theta(x,y_1) - f_\theta(x,y_2))}}{1+e^{-(f_\theta(x,y_1) - f_\theta(x,y_2))}} p_X(x) 2q(y_1, y_2|x) \mathbb{P}(y_1 \succ y_2 | x) dy_1 \tilde{f}(x,y_2)  dy_2 dx .
\end{align*}
As the functional derivative $\frac{\delta \mathcal{L}}{\delta f}$ is defined by equation $\delta \mathcal{L}[f, \tilde{f}] =  \iint \frac{\delta \mathcal{L}}{\delta f}(x,y) \tilde{f}(x,y) dy dx$, it follows that
\begin{align} 
    \frac{\delta\mathcal{L}}{\delta f_{\theta}}(x,y) 
    &= - \int \frac{e^{-(f_\theta(x,y) - f_\theta(x,y_2))}}{1+e^{-(f_\theta(x,y) - f_\theta(x,y_2))}} p_X(x) 2q(y,y_2|x) \mathbb{P}(y \succ y_2 |x) dy_2 \nonumber \\
    &\quad + \int \frac{e^{-(f_\theta(x,y_1) - f_\theta(x,y))}}{1+e^{-(f_\theta(x,y_1) - f_\theta(x,y))}} p_X(x) 2q(y_1,y|x) \mathbb{P}(y \prec y_1 |x) dy_1 \nonumber \\
    &= - \int \left( 1-\frac{e^{-(f_\theta(x,y_2) - f_\theta(x,y))}}{1+e^{-(f_\theta(x,y_2) - f_\theta(x,y))}} \right) p_X(x) 2q(y,y_2|x) \mathbb{P}(y \succ y_2 |x)  dy_2 \nonumber \\
    &\quad +  \int \frac{e^{-(f_\theta(x,y_1) - f_\theta(x,y))}}{1+e^{-(f_\theta(x,y_1) - f_\theta(x,y))}} p_X(x) 2q(y,y_1|x) \mathbb{P}(y \prec y_1 |x) dy_1 \nonumber \\
    &= - \int  p_X(x) 2q(y,y_2|x) \mathbb{P}(y \succ y_2 |x) dy_2  
    + \int \frac{e^{-(f_\theta(x,y_2) - f_\theta(x,y))}}{1+e^{-(f_\theta(x,y_2) - f_\theta(x,y))}} p_X(x) 2q(y,y_2|x) \mathbb{P}(y \succ y_2 |x)  dy_2 \nonumber \\
    &\quad +  \int \frac{e^{-(f_\theta(x,y_2) - f_\theta(x,y))}}{1+e^{-(f_\theta(x,y_2) - f_\theta(x,y))}} p_X(x) 2q(y,y_2|x) \mathbb{P}(y \prec y_2 |x) dy_2 \label{ftlderiavtive1_step1} \\
    &= - \int  p_X(x) 2q(y,y_2|x) \mathbb{P}(y \succ y_2 |x) dy_2  
    +  \int \frac{e^{-(f_\theta(x,y_2) - f_\theta(x,y))}}{1+e^{-(f_\theta(x,y_2) - f_\theta(x,y))}} p_X(x) 2q(y,y_2|x)  dy_2 . \label{ftlderivative1_step2}
\end{align}
In \eqref{ftlderiavtive1_step1}, dummy variable $y_1$ in the last integral is substituted by $y_2$. \eqref{ftlderivative1_step2} uses $\mathbb{P}(y \succ y_2|x) + \mathbb{P}(y \prec y_2|x) = 1$.

$\frac{\delta \mathcal{L}}{\delta f_\theta}$ can be further reduced by realizing that \eqref{ftlderivative1_step2} is an expectation with respect to $y_2$ over a conditional distribution as follows. \eqref{ftlderivative1_step2} is integrated by $y_2$ with the term $p_X(x) q(y,y_2|x)$. If $(X,Y_1,Y_2) \sim \mathcal{D}_u$, then this term is the joint density. However, $x$ and $y$ are arguments of $\frac{\delta \mathcal{L}}{\delta f_\theta}$, and the only variable that is being integrated is $y_2$. In order to simplify the integrals, we define $\mathcal{D}_u | (X,Y_1) = (x,y)$ as the conditional distribution of $Y_2$ given $X$ and $Y_1$ under $\mathcal{D}_u$. Equivalently, the density of $Y_2 \sim \mathcal{D}_u | (X,Y_1) = (x,y)$ is $\frac{p_X(x) q(y,y_2|x)}{p_{X,Y_1}(x,y)}$, where $p_{X,Y_1}(x,y) = \int p_X(x) q(y,y_2|x) dy_2$ is the marginal distribution of $(X,Y_1)$ on $\mathcal{D}_u$. Intuitively, it amounts to first sampling $(X,Y_1,Y_2) \sim \mathcal{D}_u$ and then choosing only the case where $(X,Y_1) = (x,y)$. With this definition, the first integral in \eqref{ftlderivative1_step2} is
\begin{align} \label{ftlderivative1_step3}
    \int  p_X(x) 2q(y,y_2|x) \mathbb{P}(y \succ y_2 |x) dy_2 &= \mathbb{E}_{Y_2 \sim \mathcal{D}_u|{(X,Y_1)=(x,y)}}[\mathbb{P}(y \sim Y_2 | x)].
\end{align}

The second integral in \eqref{ftlderivative1_step2} is reduced using the same distribution. Recall that the BT model with a reward $r$ is defined as 
\begin{align*}
    \mathbb{P}^{BT}(y_1 \succ y_2|x) = \frac{\exp(r(x, y_1))}{\exp(r(x, y_1)) + \exp(r(x, y_2))} = \sigma(r(x,y_1) - r(x,y_2)).
\end{align*}
Let us denote by $\mathbb{P}^{BT}_\theta$ the BT model with reward $ f_\theta$, that is,
\begin{align*}
    \mathbb{P}^{BT}_\theta(y_1 \succ y_2|x) = \sigma( f_\theta(x,y_1) - f_\theta(x,y_2)).
\end{align*}
Then
\begin{align}
    \int \frac{e^{-(f_\theta(x,y_2) - f_\theta(x,y))}}{1+e^{-(f_\theta(x,y_2) - f_\theta(x,y))}} p_X(x) 2q(y,y_2|x)  dy_2 
    &= \int \sigma( (f_\theta(x,y) - f_\theta(x,y_2)))  p_X(x) 2q(y,y_2|x)  dy_2 \nonumber\\
    &= \int \mathbb{P}^{BT}_\theta(y \succ y_2 |x) p_X(x) 2q (y,y_2|x) dy_2 \nonumber \\
    &= \mathbb{E}_{Y_2 \sim \mathcal{D}_u| (X,Y_1)=(x,y)}[\mathbb{P}^{BT}_\theta (y \succ Y_2|x) | x,y] p_{X,Y_1}(x,y). \label{ftlderivative1_step4}
\end{align}
Plugging \eqref{ftlderivative1_step3} and \eqref{ftlderivative1_step4} in \eqref{ftlderivative1_step2} leads to
\begin{align}
    \frac{\delta\mathcal{L}}{\delta f_\theta} (x,y) &= - \mathbb{E}_{Y_2 \sim \mathcal{D}_u |X,Y_1} \left[ \mathbb{P}(y \succ Y_2 |x) - \mathbb{P}^{BT}_\theta(y \succ Y_2|x) | x,y\right] 2p_{X,Y_1}(x,y), \label{eq:ftl_derivative}
\end{align}
and accordingly,
\begin{align*}
    g_\theta (x,y) &= - \mathbb{E}_{Y_2 \sim \mathcal{D}_u |X,Y_1} \left[ \mathbb{P}(y \succ Y_2 |x) - \mathbb{P}^{BT}_\theta(y \succ Y_2|x) | x,y\right] 2p_{X,Y_1}(x,y) \frac{\partial f_\theta}{\partial \theta}(x,y)^2 .
\end{align*}

Therefore, $g_\theta(x,y)<0$ if and only if 
\begin{align*}
    \mathbb{E}_{Y_2 \sim \mathcal{D}_u | (X,Y_1)=(x,y)} [\mathbb{P}(y \succ Y_2 | x)] 
    > \mathbb{E}_{Y_2 \sim \mathcal{D}_u | (X,Y_1)=(x,y)}\mathbb{P}^{BT}_\theta (y \succ Y_2 | x)]
\end{align*}
under which condition $f_{\theta_{t+1}}(x,y) > f_{\theta_t}(x,y)$ holds up to a negligible $O(\alpha^2)$ term.
\end{proof}

\subsubsection{Statement and Proof of Theorem \ref{thm:empirical_GD}}

\begin{theorem}\label{thm:empirical_GD}
Let $\pi_{\theta_{t+1}}$ be the policy after one step of gradient descent on the empirical DPO loss 
at step $t$. Then the updated policy is compared with the previous policy as 
\begin{enumerate}[i)]
    \item $\pi_{\theta_{t+1}}(y|x) >  \pi_{\theta_t}(y|x) \quad \text{if} \quad \hat{\mathbb{P}}_w(y|x)  > \hat{\mathbb{P}}_{w,\theta_t}(y|x)$
    \item $ \pi_{\theta_{t+1}}(y|x) =  \pi_{\theta_t}(y|x) \quad \text{if} \quad \hat{\mathbb{P}}_w(y|x)  = \hat{\mathbb{P}}_{w,\theta_t}(y|x) \text{ or } C=\phi$
    \item $\pi_{\theta_{t+1}}(y|x) <  \pi_{\theta_t}(y|x) \quad \text{if} \quad \hat{\mathbb{P}}_w(y|x)  < \hat{\mathbb{P}}_{w,\theta_t}(y|x)$.
\end{enumerate}
\end{theorem}

\begin{proof}
Adopting the proof of Theorem \ref{thm:gradiennt},
\begin{align*}
    f_{\theta_{t+1}}(x,y) = f_{\theta_t}(x,y) - \alpha g_{\theta_t}(x,y) + O(\alpha^2)
\end{align*}
from \eqref{eq:policy_GD}, where $g_\theta(x,y) = \frac{\delta \hat{\mathcal{L}}}{\delta f_\theta}(x,y) \frac{\partial f_\theta}{\partial \theta} (x,y)^2$. Also, \eqref{ftldifferential} extends to
\begin{align*}
    \delta \hat{\mathcal{L}} [f,\tilde{f}] 
    &= -\frac{1}{n} \sum_{i=1}^n \frac{e^{-(f_\theta(x_i,y_{wi}) - f_\theta(x_i,y_{li}))}}{1+e^{-(f_\theta(x_i,y_{wi}) - f_\theta(x_i,y_{li}))}} (\tilde{f}(x_i,y_{wi}) - \tilde{f}(x_i,y_{li}))
\end{align*}
where we denote the dataset as $\{(x_i y_{wi}, y_{li}) | i=1, \cdots, n\}$. 

By the relation $\delta \hat{\mathcal{L}}[f, \tilde{f}] =  \iint \frac{\delta \hat{\mathcal{L}}}{\delta f}(x,y) \tilde{f}(x,y) dx dy$, it follows that
\begin{align} \label{ftlderivative}
    \frac{\delta\hat{\mathcal{L}}}{\delta f_{\theta}}(x,y) 
    = -\frac{1}{n} \sum_{i=1}^n \frac{e^{-(f_\theta(x,y) - f_\theta(x,y_{li}))}}{1+e^{-(f_\theta(x,y) - f_\theta(x,y_{li}))}} \delta_{x_i}(x) \delta_{y_{wi}}(y) 
    + \frac{1}{n} \sum_{i=1}^n \frac{e^{-(f_\theta(x,y_{wi}) - f_\theta(x,y))}}{1+e^{-(f_\theta(x,y_{wi}) - f_\theta(x,y))}} \delta_{x_i}(x) \delta_{y_{li}}(y)
\end{align}
where $\delta_c (\cdot) = \delta(\cdot-c)$ is the time-delayed Dirac delta function. Note that \eqref{ftlderivative} is a Schwartz distribution and holds only in the weak sense. Identifying \eqref{ftlderivative} with a function by replacing $\delta$ with the Dirac measure, we have
\begin{align*} 
    \frac{\delta\hat{\mathcal{L}}}{\delta f_{\theta}}(x,y) 
    = -\frac{1}{n} \sum_{i=1}^n \frac{e^{-(f_\theta(x_i,y_{wi}) - f_\theta(x_i,y_{li}))}}{1+e^{-(f_\theta(x_i,{y_{wi}}) - f_\theta(x_i,y_{li}))}} \left( \bm{1}\{(x,y)=(x_i,y_{wi})\} - \bm{1}\{(x,y)=(x_i,y_{li})\} \right)
\end{align*}
which is defined everywhere. Further simplifying the expression, 
\begin{align}
    \frac{\delta\hat{\mathcal{L}}}{\delta f_{\theta}}(x,y) 
    &= -\frac{1}{n} \sum_{i:(x_i,y_{wi})=(x,y)}  \frac{e^{-(f_\theta(x,y) - f_\theta(x,y_{li}))}}{1+e^{-(f_\theta({x,y}) - f_\theta(x,y_{li}))}}
    +\frac{1}{n} \sum_{i:(x_i,y_{li})=(x,y)}  \frac{e^{-(f_\theta(x,y_{wi}) - f_\theta(x,y))}}{1+e^{-(f_\theta(x,{y_{wi}}) - f_\theta(x,y))}} \nonumber \\
    &= -\frac{1}{n} \sum_{i:(x_i,y_{wi})=(x,y)} \left( 1- \frac{e^{-(f_\theta(x,y_{li}) - f_\theta(x,y))}}{1+e^{-(f_\theta(x,{y_{li}}) - f_\theta(x,y))}} \right)
    +\frac{1}{n} \sum_{i:(x_i,y_{li})=(x,y)}  \frac{e^{-(f_\theta(x,y_{wi}) - f_\theta(x,y))}}{1+e^{-(f_\theta(x,{y_{wi}}) - f_\theta(x,y))}} \nonumber \\
    &= -\frac{1}{n} \sum_{i:(x_i,y_{wi})=(x,y)}1 + \frac{1}{n} \sum_{z :(x,y,z) \in \mathcal{D} \text{ or } (x,z,y) \in \mathcal{D}} \sigma(f_\theta(x,y) - f_\theta(x,z)) \label{ftlderivative2_step1}.
\end{align}

Computing $g_\theta$ at a point $(x,y)$ requires taking sum over the samples with $x_i = x$ and $y_{wi}=y$ or $y_{li}=y$. 
Let $C$ be the set of responses that competes with $y$ for a prompt $x$, that is, 
\begin{align*}
    C = \{y_{li} | x_i=x, y_{wi}=y\} \cup \{y_{wi} | x_i=x, y_{li}=y\}.
\end{align*} 
Recalling that $\sigma(f_\theta(x,y) - f_\theta(x,z))$ is the BT model with the reward being $r(x,y) = f_\theta (x,y)$, \eqref{ftlderivative2_step1} becomes
\begin{align} \label{ftlderivative2}
    g_{\theta}(x,y) &= \frac{1}{n} \left\{-\sum_{\tilde{y} \in C} \bm{1}\{y \succ \tilde{y} \} +  \sum_{\tilde{y} \in C} \mathbb{P}^{BT}_\theta( y \succ \tilde{y} | x)\right\} \frac{\partial f_\theta}
    {\partial \theta}(x,y)^2.
\end{align} 
Therefore, $g_\theta(x,y)<0$ if and only if 
\begin{align} \label{ftlderivative2_comparison}
    \sum_{\tilde{y} \in C} \bm{1}\{y \succ \tilde{y} \}
    > \sum_{\tilde{y} \in C} \mathbb{P}^{BT}_\theta( y \succ \tilde{y} | x)
\end{align}
under which condition $f_{\theta_{t+1}}(x,y) > f_{\theta_t}(x,y)$ holds up to a negligible $O(\alpha^2)$ term. 

\eqref{ftlderivative2_comparison} provides intuition to how the gradient descent on the empirical loss changes the policy. After dividing both sides of \eqref{ftlderivative2_comparison} with $|C|$, provided that $C \neq \phi$, the left hand side becomes the proportion of samples that $y$ is actually the preferred response among the samples that $y$ is a response. The right hand side is the average prediction of BT model that $y$ will be the preferred response. On the given dataset, if the frequency of $y$ being the preferred response is greater than the prediction of the BT model, then it implies that the current model is generating $y$ with a lower probability than it is preferred on the actual dataset. In this case, gradient descent on DPO increases the likelihood of $y$. Also, if $C = \phi$, which happens if $(x,y)$ never appears as a prompt-response pair in the dataset, then no information can be obtained. Accordingly, $g_\theta(x,y) = 0$ and the policy is not updated at $(x,y)$.  

\eqref{ftlderivative2_comparison} also explains the magnitude of the update. The change in the log-probability $f_{\theta_{t_1}}(x,y) - f_{\theta_t}(x,y)$ is dominated by the gradient. By \eqref{ftlderivative2}, the change is proportional to the gap $\sum_{\tilde{y} \in C} \bm{1}\{y \succ \tilde{y} \} - \sum_{\tilde{y} \in C} \mathbb{P}^{BT}_\theta( y \succ \tilde{y} | x)$. Therefore, if the difference between the actual and the predicted probability of $y$ being the preferred response is larger, then the policy update at $y$ is greater.
\end{proof}

\subsection{Proofs Related to Section \ref{section:linear}}

In this section, Lemma \ref{lem:linear_RLHF} and Proposition \ref{prop_linear_RLHF} are proved.

\subsubsection{Proof of Lemma \ref{lem:linear_RLHF}}

\begin{replemma}{lem:linear_RLHF}
(restated) Starting from some Gaussian reference policy $\pi_{\theta_{\text{ref}}}$ with $\theta_{\text{ref}} = (w_{\text{ref}}, \sigma_{\text{ref}})$, the policy $\pi_{\theta^*}$ that minimizes RLHF loss \eqref{RLHF_original} follows
\begin{align*}
    \pi_{\theta^*}(\cdot|x) \sim \mathcal{N}\left( (\gamma w_{\text{ref}} + (1-\gamma)w^*)^\top x\,, \,\,\frac{\sigma_{\text{ref}}^2\beta}{\beta+2\sigma_{\text{ref}}^2}\right), \,\,\text{ where } \gamma =\frac{\beta}{\beta+2\sigma_{\text{ref}}^2}.
\end{align*}
Namely, $\pi_{\theta^*}$ is also a Gaussian policy.
\end{replemma}

\begin{proof}
From Proposition \ref{prop_optimal_loss}, the minimizer of \eqref{RLHF_original} is
\begin{align}\label{lemma1_sol}
    \pi_{\theta^*}(y|x) &=  \frac{1}{Z(x)}\pi_{\theta_{\text{ref}}}(y|x)\exp\left(\frac{1}{\beta}r^*(x,y)\right)
\end{align}
where $Z(x) = \int \pi_{\text{ref}}(y|x) \exp\left(\frac{1}{\beta}r^*(x,y)\right) dy$.
The model family we consider is 
\begin{align*}
    \pi_\theta(y|x) = \frac{1}{\sqrt{2\pi\sigma^2}}\exp\left(-\frac{(y-w^\top x)^2}{2\sigma^2}\right)
\end{align*}
with reward $r^*(x,y) = -(y-w^{*\top}x)^2$. Plugging in the reference policy $\pi_{\text{ref}}$ parametrized by $(w_{\text{ref}}, \sigma_{\text{ref}}^2)$ and the reward $r^*$ to \eqref{lemma1_sol} gives
\begin{align*}
    \pi_{\theta^*}(y|x) &= \frac{1}{Z(x)} \frac{1}{\sqrt{2\pi\sigma_{\text{ref}}^2}} \exp\left(-\frac{(y-w_{\text{ref}}^\top x)^2}{2\sigma_{\text{ref}}^2}\right) \exp\left(-\frac{1}{\beta}(y-w^{*\top}x)^2\right) \\
    &\propto \exp\left( -\left(\frac{1}{2\sigma_{\text{ref}}^2}+\frac{1}{\beta}\right)y^2 + \left(\frac{w_{\text{ref}}}{\sigma_{\text{ref}}^2}+\frac{2w^*}{\beta}\right)^\top x y \right) \\
    &\propto \exp\left( -\left(\frac{1}{2\sigma_{\text{ref}}^2}+\frac{1}{\beta}\right) \left(y-\frac{\left(\frac{w_{\text{ref}}}{2\sigma_{\text{ref}}^2}+\frac{w^*}{\beta}\right)^\top x}{\frac{1}{2\sigma_{\text{ref}}^2}+\frac{1}{\beta}}\right)^2 \right).
\end{align*}
Therefore,
\begin{align*}
    \pi_{\theta^*}(\cdot|x) &= \mathcal{N}\left( \frac{\left(\frac{w_{\text{ref}}}{2\sigma_{\text{ref}}^2}+\frac{w^*}{\beta}\right)^\top x}{\frac{1}{2\sigma_{\text{ref}}^2}+\frac{1}{\beta}}, \frac{1}{2\left(\frac{1}{2\sigma_{\text{ref}}^2}+\frac{1}{\beta}\right)}\right)\\
    &= \mathcal{N}\left( \left\{\gamma w_{\text{ref}} + (1-\gamma)w^*\right\}^\top x, \frac{\sigma_{\text{ref}}^2\beta}{\beta+2\sigma_{\text{ref}}^2}\right)
\end{align*}
where $\gamma = \frac{\beta}{\beta + 2\sigma_{\text{ref}}^2}$.
\end{proof}

\subsubsection{Proof of Proposition \ref{prop_linear_RLHF}}

\begin{repproposition}{prop_linear_RLHF}
(restated) Consider the linear alignment model, and apply Algorithm~\ref{alg:online_DPO} with $\theta_0 = (w_0, \sigma_0)$, number of iterations $T$, and data generation procedure $\text{ALG}_{\text{Data}}$ satisfying Assumption~\ref{assumption:linear_RLHF}. At iteration $t$, the minimizing policy $ \pi_{\theta_t}(\cdot \mid x) \sim \mathcal{N}(w_t^\top x, \sigma_t^2)$, where
\begin{align*}
    w_t= w^* + \frac{\beta}{\beta+2t\sigma_0^2}(w_0-w^*) \quad \text{and} \quad \sigma_t^2 = \frac{\beta \sigma_0^2}{\beta + 2t \sigma_0^2}.
\end{align*}
Moreover, the generated response $Y_t \sim \pi_{\theta_t}(\cdot \mid x)$ converges to $w^{*\top} x$ in probability as $t \to \infty$.
\end{repproposition}

\begin{proof}
Applying Lemma \ref{lem:linear_RLHF} with $\pi_{\theta_t}$ as reference policy leads to the optimal policy for the RLHF loss \eqref{RLHF_original} as 
\begin{align*}
    \pi_{\theta_{t+1}}(\cdot|x) = \mathcal{N}\left(\left\{\gamma w_t + (1-\gamma) w^*\right\}^\top x, \frac{\sigma_t^2\beta}{\beta+2\sigma_t^2} \right)
\end{align*}
where $\gamma_t = \frac{\beta}{\beta+2\sigma_t^2}$. By Proposition \ref{prop_optimal_loss}, this is also an optimal policy for the DPO loss \eqref{DPO_form_f} on $supp(\mathcal{D}_{Y|x})$. Since $y$ can take any value on $\mathbb{R}$ by Assumption \ref{assumption:linear_RLHF}, this is optimal policy for the DPO loss everywhere.
Hence
\begin{align*}
    w_{t+1} = \gamma_t w_t + (1-\gamma_t) w^* \quad \text{and} \quad \sigma_{t+1}^2 = \frac{\sigma_t^2 \beta}{\beta + 2\sigma_t^2}.
\end{align*}
Taking reciprocal of $\sigma_{t+1}^2$ gives $\frac{1}{\sigma_{t+1}^2} = \frac{1}{\sigma_t^2}+\frac{2}{\beta}$ for all $t$. Therefore, $\frac{1}{\sigma_t^2} = \frac{1}{\sigma_0^2} + \frac{2t}{\beta}$, or equivalently,
\begin{align} \label{prop_linear_RLHF_sigma}
    \sigma_t^2 = \frac{\beta\sigma_0^2}{\beta+2t\sigma_0^2}.
\end{align}
Rearranging the terms of $w_{t+1}$ gives $w_{t+1} -w^* = \gamma_t (w_t-w^*)$ for all $t$. Recursively applying the relation leads to $w_t-w^* = \Pi_{k=0}^{t-1}\gamma_k (w_0-w^*)$. Using \eqref{prop_linear_RLHF_sigma}, $\gamma_t = \frac{\beta}{\beta+2\sigma_t^2}$ reduces to $\gamma_k = \frac{\beta + 2k\sigma_0^2}{\beta + 2(k+1)\sigma_0^2}$. Therefore,
\begin{align} \label{prop_linear_RLHF_w}
    w_t-w^* = \frac{\beta}{\beta+2t\sigma_0^2} (w_0-w^*).
\end{align}
Now, from \eqref{prop_linear_RLHF_w} and \eqref{prop_linear_RLHF_sigma},
\begin{align*}
    w_t \rightarrow w^* \quad \text{and} \quad \sigma_t \rightarrow 0 \quad \text{as} \quad t\rightarrow \infty.
\end{align*}
For $Y_t \sim \mathcal{N}(w_t^\top x, \sigma_t^2)$ and $z\in \mathbb{R}$, the cumulative probability at $z$ is $\mathbb{P}(Y_t\leq z) = \Phi\left(\frac{z-w_t^\top x}{\sigma_t^2}\right)$. Hence $\mathbb{P}(Y_t\leq z)\rightarrow 0$ if $z<w^{*\top}x$ and $\mathbb{P}(Y_t\leq z)\rightarrow 1$ if $z>w^{*\top}x$ as $t \rightarrow \infty$. Therefore, $Y_t \overset{d}{\rightarrow} w^{*\top}x$, which finally leads to $Y_t \overset{p}{\rightarrow} w^{*\top}x$.
\end{proof}

\subsection{Proofs Related to Section \ref{section:convergence}}

In this section, Lemma \ref{lem:analytical}, Proposition \ref{prop_fisher}, Observation \ref{observation:approximation}, and Observation \ref{observation:approximation2} are proved.

\subsubsection{Proof of Lemma \ref{lem:analytical}}

\begin{replemma}{lem:analytical}
(restated) For a given $x$, $\mathbb{E}[\epsilon_1^2]$ and $\mathbb{E}[|\epsilon_1 - \epsilon_2|]$ are functions of $K$ and $\delta(x)$ such that
\begin{align*}
    \eta(K, \delta(x)) &:= \mathbb{E}[\epsilon_1^2] = K\int z^2\varphi(z) \left(1-F_{|\delta(x)+Z|}(|\delta(x)+z|)\right)^{K-1}dz\\
    \gamma(K, \delta(x)) &:= \mathbb{E}[|\epsilon_1-\epsilon_2|] = K\int \varphi(z) \left(1-F_{|\delta(x)+Z|}(|\delta(x)+z|)\right)^{K-1}\left(z(2\Phi(z) - 1) + 2\varphi(z) \right)dz,
\end{align*}
where $\varphi$ and $\Phi$ denote the standard normal density and CDF, respectively, and $F_{|\delta(x)+Z|}(\cdot)$ is the CDF of the random variable $|\delta(x) + Z|$ with $Z \sim \mathcal{N}(0,1)$.
\end{replemma}
\begin{proof}
Let us denote the PDF and CDF of the standard normal distribution as $\phi$ and $\Phi$, respectively. The two noises $\epsilon_1$ and $\epsilon_2$ are independent, where $\epsilon_1$ is generated by Steps \ref{alg_line:sampling_start}–\ref{alg_line:sampling_end} of Algorithm \ref{alg:Data} and $\epsilon_2$ follows the standard normal distribution. The distribution of $\epsilon_1$ is 
\begin{align*}
    p_{\epsilon_1}(u) = K \phi(u) \left\{1-F_{|\delta(x)+Z|}(|\delta(x)+u|) \right\}^{K-1}
\end{align*}
by Lemma \ref{lemma:resample_pdf}. Therefore,  
\begin{align}
    \gamma(K, \delta(x))& := \mathbb{E}\left[ | \epsilon_1 - \epsilon_2 | \right] \nonumber \\
    &= \iint |\epsilon_1 - \epsilon_2| K \phi(\epsilon_1) \left\{1-F_{|\delta(x)+Z|}(|\delta(x)+\epsilon_1|)\right\}^{K-1} \phi(\epsilon_2) d\epsilon_2 d\epsilon_1 \nonumber \\
    &= \int K\phi(\epsilon_1) \left\{1-F_{|\delta(x)+Z|}(|\delta(x)+\epsilon_1|)\right\}^{K-1} \left(\epsilon_1 (2\Phi(\epsilon_1)-1) + 2\phi(\epsilon_1) \right) d\epsilon_1 \label{eq:mean_final},
\end{align}
where \eqref{eq:mean_final} follows from
\begin{align*}
    \int |\xi-a|\phi(\xi) d\xi &= \int_{-\infty}^a (a-\xi) \frac{1}{\sqrt{2\pi}}e^{-\frac{\xi^2}{2}} d\xi + \int_a^\infty (\xi-a) \frac{1}{\sqrt{2\pi}}e^{-\frac{\xi^2}{2}} d\xi \\
    &= a \Phi(a) + \left[\frac{1}{\sqrt{2\pi}}e^{-\frac{\xi^2}{2}}\right]_{-\infty}^a + \left[-\frac{1}{\sqrt{2\pi}}e^{-\frac{\xi^2}{2}}\right]_a^\infty - a (1-\Phi(a)) \\
    &= a(2\Phi(a)-1) + 2\phi(a).
\end{align*}
In addition,
\begin{align}
    \eta(K, \delta(x)) &:= \mathbb{E}[\epsilon_1^2] \nonumber \\
    &= \int \epsilon_1^2 K \phi(\epsilon_1) \left\{1-F_{|\delta(x)+Z|}(|\delta(x)+\epsilon_1|)\right\}^{K-1} d\epsilon_1 \label{eq:var_final} 
\end{align}
Replacing the dummy variable with $z$ in \eqref{eq:mean_final} and \eqref{eq:var_final} concludes the proof.
\end{proof}

\subsubsection{Proof of Proposition \ref{prop_fisher}}

\begin{repproposition}{prop_fisher}
(restated) Let $N$ denote the sample size of $\mathcal{D}$, with $\{x_n\}_{n=1}^N$ representing the current sampled inputs, and let $t$ denote the current iteration of the online DPO training. For the current parameter $w_t$, the magnitude of the gradient is bounded by
\begin{align*}
\lVert\nabla_w\mathcal{L}(w_t)\rVert \leq \begin{cases}
    \frac{\beta}{2N\sigma_{t}}\sum_{n=1}^N \gamma(K,\delta(x_n)) \Vert x_n \rVert  & K > 1\\
    \frac{\beta}{2N\sigma_{t}} \frac{1}{\sqrt{\pi}}\sum_{n=1}^N  \Vert x_n \rVert  & K = 1
\end{cases} 
\end{align*}
and the Fisher information matrix is
\begin{align*}
I_K(w_t) = \begin{cases}
    \frac{\beta^2}{4N\sigma_{t}^2} \sum_{n=1}^N \left\{ \eta(K, \delta(x_n)) +1 \right\} x_n x_n^\top & K > 1\\
    \frac{\beta^2}{2N\sigma_{t}^2} \sum_{n=1}^N x_n x_n^\top  & K = 1
\end{cases} 
\end{align*}
\end{repproposition}

\begin{proof}
In the online DPO setup implemented by Algorithm \ref{alg:online_DPO} equipped with Algorithm \ref{alg:Data} as the data generating algorithm, the current policy $\pi_{\theta_t}$ serves both as the reference policy and also the data generating distribution. Responses $y_1$ and $y_2$ are generated from the reference policy $\pi_{ref} (=\pi_{\theta_t})$ by either standard sampling or best-of-$K$ sampling. The new policy $\pi_{\theta_{t+1}}$ is obtained by updating the old policy $\pi_{\theta_t}$ with the gradient of the loss evaluated at $\theta_t$. 

As in Lemma \ref{lemma:DPO_loss_grad}, let us denote by $(x,y_+,y_-)$ the preference data and $\tilde{\sigma}$ (to distinguish it from the standard deviation $\sigma$ of the Gaussian model) the logistic function. Recall that the loss on the dataset $\{(x_n, y_{+n}, y_{-n})\}_{n=1}^N$ is
\begin{align*}
    \hat{\mathcal{L}}(\theta; \{(x_n, y_{+n}, y_{-n})\}_{n=1}^N) 
    &= -\frac{1}{N}\sum_{n=1}^N \log{\tilde{\sigma}(f_\theta(x_n,y_{+n}) - f_\theta(x_n,y_{-n}))} \\
    &= \frac{1}{N}\sum_{n=1}^n \ell(\theta; x_n, y_{+n}, y_{-n}) 
\end{align*}
and $\mathcal{L}(\theta,\mathcal{D}) = \mathbb{E}_{(x_n, y_{+n}, y_{-n}) \overset{iid}{\sim} \mathcal{D}}[\hat{\mathcal{L}}(\theta; \{(x_n, y_{+n}, y_{-n})\}_{n=1}^N)]$.

Two key observations are crucial for analyzing the update. First, at $\theta=\theta_{ref}$, the function $f_\theta(x,y)$ takes the value $0$. Second, the pair $(\epsilon_+,\epsilon_-)$ is either $(\epsilon_1,\epsilon_2)$ or $(\epsilon_2,\epsilon_1)$, where $\epsilon_1 = \frac{y_1-w_{ref}^\top x}{\sigma_{ref}}$, $\epsilon_2=\frac{y_2-w_{ref}^\top x}{\sigma_{ref}}$, and $\epsilon_+$ and $\epsilon_-$ are defined in Lemma \ref{lemma:DPO_loss_grad}. Combining these two properties along with the gradient \eqref{eq:grad} and the Hessian \eqref{eq:grad3} in Lemma \ref{lemma:DPO_loss_grad} gives
\begin{align*}
    \left\lVert \left.\frac{\partial l}{\partial w} \right|_{\theta=\theta_{ref}}\right\rVert 
    &= \beta\frac{1}{\sigma_{ref}} \frac{1}{2} |\epsilon_+ - \epsilon_-| \lVert x \rVert \\
    &= \frac{\beta}{2\sigma_{ref}}  |\epsilon_1 - \epsilon_2| \lVert x \rVert
\end{align*}
and 
\begin{align*}
    \left.\frac{\partial^2 l}{\partial w^2} \right|_{\theta=\theta_{ref}} 
    &=\beta^2 \frac{1}{\sigma_{ref}^2} \frac{1}{4} (\epsilon_+-\epsilon_-)^2 xx^\top \\
    &= \frac{\beta^2}{4\sigma_{ref}^2}(\epsilon_1-\epsilon_2)^2 xx^\top.
\end{align*}
Since the reference policy is the old policy in online DPO, $\theta_{ref} $ is $ \theta_t$, or equivalently, $(w_{ref}, \sigma_{ref}) = (w_t,\sigma_t)$.
Therefore, given $x_1, \cdots, x_N$,
\begin{align*}
    \lVert \nabla_w \mathbb{E}[ \hat{\mathcal{L}} |x_1, \cdots, x_N ] (w_t) \rVert 
    &= \left\lVert \frac{1}{N} \sum_{n=1}^N \mathbb{E}[\nabla_w \ell(w_t)|x_n] \right\rVert \\
    &\leq \frac{1}{N}\sum_{n=1}^N \left\lVert \mathbb{E}\left[ \nabla_w \ell(w_t) |x_n \right] \right\rVert \\
    &\leq \frac{1}{N}\sum_{n=1}^N \mathbb{E}\left[ \left\lVert  \nabla_w \ell(w_t)  \right\rVert |x_n\right] \\
    &= \frac{\beta}{2N\sigma_{t}}\sum_{n=1}^N \mathbb{E}\left[ |\epsilon_{1n}-\epsilon_{2n} | \Vert x_n \rVert |x_n \right]\\
    &= \frac{\beta}{2N\sigma_{t}}\sum_{n=1}^N \gamma(K,\delta(x_n)) \Vert x_n \rVert \\
\end{align*}
and
\begin{align*}
    \nabla_w^2 \mathbb{E}[\hat{\mathcal{L}}|x_1, \cdots, x_N] (w_t) 
    &= \frac{1}{N} \sum_{n=1}^N \mathbb{E}\left[ \nabla_w^2\ell(w_t) |x_n\right] \\
    &= \frac{\beta^2}{4N\sigma_{t}^2} \sum_{n=1}^N \mathbb{E}\left[(\epsilon_{1n}-\epsilon_{2n})^2  x_n x_n^\top |x_n\right]\\
    &= \frac{\beta^2}{4N\sigma_{t}^2} \sum_{n=1}^N \mathbb{E}\left[(\epsilon_{1n}^2+\epsilon_{2n}^2-2\epsilon_{1n}\epsilon_{2n})  x_n x_n^\top |x_n\right]\\
    &= \frac{\beta^2}{4N\sigma_{t}^2} \sum_{n=1}^N \left\{ \eta(K, \delta(x_n)) +1 \right\} x_n x_n^\top \\
\end{align*}
since $\epsilon_{2n} \sim \mathcal{N}(0,1)$ and $\epsilon_{2n}$ is independent of $\epsilon_{1n}$ for all $n$. 

The case where $K=1$ corresponds to the standard sampling, in which circumstance $\epsilon_1$ and $\epsilon_2$ are independent standard Gaussian noises. Hence $\gamma(1,\delta(x)) = \mathbb{E}[|\epsilon_1 - \epsilon_2|] = \frac{1}{\sqrt{\pi}}$ since $\epsilon_1 - \epsilon_2 \sim \mathcal{N}(0,2)$. Also, $\eta(1,\delta(x)) = \mathbb{E}[\epsilon_1^2] = 1$.

\end{proof}

\subsubsection{Proof of Observation \ref{observation:approximation}}

\begin{repobservation}{observation:approximation}
(restated) Let $\gamma(K,\delta(x))$ and $\eta(K,\delta(x))$ be defined as in Lemma \ref{lem:analytical}. For large enough $|\delta(x)|$, the order of the two functions are
\begin{align*}
    \gamma(K,\delta(x)) = O(|\delta(x)|) \quad \text{and} \quad \eta(K,\delta(x)) = O(\delta(x)^2).
\end{align*}
\end{repobservation}

\begin{proof}
We prove the case where $\delta(x)>0$. The case where $\delta(x)<0$ can be proved in the same way.
From Lemma \ref{lem:analytical},
\begin{align*}
    \gamma(K,\delta(x)) = K \int \phi(z) \left\{1-F_{|\delta(x)+Z|}(|\delta(x)+z|) \right\}^{K-1} \left\{z(2\Phi(z)-1)+2\phi(z)\right\}dz.
\end{align*}
Since
\begin{align*}
    F_{|\delta(x)+Z|}(|\delta(x)+z|) 
    &= \mathbb{P}(-|\delta(x)+z|\leq\delta(x)+Z\leq|\delta(x)+z|)\\
    &= \Phi(-\delta(x)+|\delta(x)+z|)-\Phi(-\delta(x)-|\delta(x)+z|) \\
    &= \begin{cases}
        \Phi(-2\delta(x)-z)-\Phi(z) & \text{if } z<-\delta(x) \\
        \Phi(z)-\Phi(-2\delta(x)-z) & \text{if } z>-\delta(x)
    \end{cases},
\end{align*}
$\gamma(K,\delta(x))$ can be reorganized as
\begin{align} \label{eq:gamma}
\begin{split}
    \gamma(K,\delta(x)) 
    &= \underbrace{K \int_{-\infty}^{-\delta(x)} \phi(z) \left\{1-\Phi(-2\delta(x)-z)+\Phi(z) \right\}^{K-1} \left\{z(2\Phi(z)-1)+2\phi(z)\right\}dz}_{\textcircled{1}} \\
    &\underbrace{\quad + K \int_{-\delta(x)}^{\infty} \phi(z) \left\{1-\Phi(z)+\Phi(-2\delta(x)-z)\right\}^{K-1} \left\{z(2\Phi(z)-1)+2\phi(z)\right\}dz}_{\textcircled{2}}.
\end{split}
\end{align}
The first term is nonnegative and is bounded by
\begin{align*}
    \textcircled{1}
    &\leq K \int_{-\infty}^{-\delta(x)} \phi(z)  \left\{z(2\Phi(z)-1)+2\phi(z)\right\}dz \\
    &\leq K \int_{-\infty}^{-\delta(x)} \phi(z)  \left\{-z+2\phi(z)\right\}dz \\
    &= O(K)
\end{align*}
for large enough $\delta(x)$.

The second term is also nonnegative and
\begin{align*}
    \textcircled{2} &\leq K \int_{-\delta(x)}^\infty \left\{\phi(z)+\phi(-2\delta(x)-z)\right\} \left\{1-\Phi(z)+\Phi(-2\delta(x)-z)\right\}^{K-1} \left\{z(2\Phi(z)-1)+2\phi(z)\right\}dz \\
    &= \left[-\left\{z(2\Phi(z)-1)+2\phi(z)\right\} \left\{1-\Phi(z)+\Phi(-2\delta(x)-z)\right\}^K \right]_{-\delta(x)}^\infty \\
    &\quad + \int_{-\delta(x)}^\infty \left\{1-\Phi(z)+\Phi(-2\delta(x)-z)\right\}^{K} \left\{2\Phi(z)-1\right\}dz \\
    &= -\delta(x)(2\Phi(-\delta(x))-1) + 2\phi(-\delta(x)) 
    + \int_{-\delta(x)}^\infty \left\{1-\Phi(z)+\Phi(-2\delta(x)-z)\right\}^{K} \left\{2\Phi(z)-1\right\}dz \\
    &\leq -\delta(x)(2\Phi(-\delta(x))-1) + 2\phi(-\delta(x)) 
    + \int_{0}^\infty \left\{1-\Phi(z)+\Phi(-2\delta(x)-z)\right\}^{K} \left\{2\Phi(z)-1\right\}dz \\
    &\leq -\delta(x)(2\Phi(-\delta(x))-1) + 2\phi(-\delta(x)) 
    + \int_{0}^1 \left\{2\Phi(z)-1\right\}dz + \int_{1}^\infty \left\{2\Phi(-z)\right\}^{K}dz \\
    &\leq -\delta(x)(2\Phi(-\delta(x))-1) + 2\phi(-\delta(x)) 
    + \int_{0}^1 \left\{2\Phi(z)-1\right\}dz + \int_{1}^\infty \left\{2\frac{\phi(z)}{z}\right\}^{K}dz \\
    &\leq -\delta(x)(2\Phi(-\delta(x))-1) + 2\phi(-\delta(x)) 
    + \int_{0}^1 \left\{2\Phi(z)-1\right\}dz +  \left(\frac{2}{\sqrt{2\pi}}\right)^K \int_{0}^\infty e^{-\frac{Kz^2}{2}}dz \\
    &=O(\delta(x)).
\end{align*}
Therefore,
\begin{align*}
    \gamma(K,\delta(x)) &= \textcircled{1}+\textcircled{2} = O(\delta(x))
\end{align*}
for large $\delta(x)$.

Similarly, $\eta(K,\delta(x))$ can be reorganized as
\begin{align} \label{eq:eta}
\begin{split}
    \eta(K,\delta(x)) &= 
    \underbrace{K \int_{-\infty}^{-\delta(x)} z^2 \phi(z) \left\{1-\Phi(-2\delta(x)-z)+\Phi(z) \right\}^{K-1} dz}_{\textcircled{3}} \\
    &\underbrace{\quad + K \int_{-\delta(x)}^{\infty} z^2 \phi(z) \left\{1-\Phi(z)+\Phi(-2\delta(x)-z)\right\}^{K-1} dz}_{\textcircled{4}}.
\end{split}
\end{align}
The first term is nonnegative and is bounded by
\begin{align*}
    \textcircled{3} &\leq \int_{-\infty}^{-\delta(x)}  K z^2 \phi(z) dz = O(K)
\end{align*}
for large $\delta(x)$.

The second term is also nonnegative and
\begin{align*}
    \textcircled{4} 
    &\leq \int_{-\delta(x)}^\infty K z^2 (\phi(z)+\phi(-2\delta(x)-z)) \left\{1-\Phi(z)+\Phi(-2\delta(x)-z)\right\}^{K-1} dz \\
    &= \left[-z^2 \left\{1-\Phi(z)+\Phi(-2\delta(x)-z)\right\}^{K} \right]_{-\delta(x)}^\infty + \int_{-\delta(x)}^\infty 2z \left\{1-\Phi(z)+\Phi(-2\delta(x)-z)\right\}^{K} dz \\
    &= \delta(x)^2 + \int_{-\delta(x)}^\infty 2z \left\{1-\Phi(z)+\Phi(-2\delta(x)-z)\right\}^{K} dz \\
    &\leq \delta(x)^2 + \int_{0}^\infty 2z \left\{1-\Phi(z)+\Phi(-2\delta(x)-z)\right\}^{K} dz \\
    &\leq \delta(x)^2 + \int_{0}^1 2z dz + \int_{1}^\infty 2z \left\{2\Phi(-z)\right\}^{K} dz \\
    &\leq \delta(x)^2 + 1 + \int_{1}^\infty 2z \left\{2\frac{\phi(z)}{z}\right\}^{K} dz \\
    &= \delta(x)^2 + 1 + \int_{1}^\infty 2\left(\frac{2}{\sqrt{2\pi}}\right)^K \frac{e^{-\frac{Kz^2}{2}}}{z^{K-1}} dz \\
    &\leq \delta(x)^2 + 1 + 2\left(\frac{2}{\sqrt{2\pi}}\right)^K \int_{0}^\infty e^{-\frac{Kz^2}{2}}dz \\
    &= O(\delta(x)^2).
\end{align*}
Therefore,
\begin{align*}
    \eta(K,\delta(x)) = O(\delta(x)^2)
\end{align*}
for large $\delta(x)$.
\end{proof}

\subsubsection{Proof of Observation \ref{observation:approximation2}}

\begin{repobservation}{observation:approximation2}
(restated) The functions $\gamma(K,\delta(x))$ and $\eta(K,\delta(x))$ in Lemma \ref{lem:analytical} are bounded as follows for small $|\delta(x)|$:
\begin{align*}
    \lim_{\delta(x) \rightarrow 0} \gamma(K,\delta(x)) \geq \sqrt{\frac{2}{\pi}}
    \quad \text{and} \quad 
    \lim_{\delta(x) \rightarrow 0} \eta(K,\delta(x)) \leq \frac{1}{2}.
    \end{align*}
\end{repobservation}

\begin{proof}
From \eqref{eq:gamma},
\begin{align*}
    \gamma(K,\delta(x)) 
    &= K \int_{-\infty}^{\infty} \phi(z)  \left\{z(2\Phi(z)-1)+2\phi(z)  \right\} \\
    &\quad \times \big\{ \left\{1-\Phi(-2\delta(x)-z)+\Phi(z) \right\}^{K-1} \bm{1}\{z<-\delta(x)\} \\
    & \quad + \left\{1-\Phi(z)+\Phi(-2\delta(x)-z)\right\}^{K-1} \bm{1}\{z>-\delta(x)\} \big\}dz.
\end{align*}
The integrand is bounded by $K\phi(z) \left\{z(2\Phi(z)-1)+2\phi(z)  \right\}$, which is integrable on $\mathbb{R}$. Also, the integrand converges to 
\begin{align*}
    K\phi(z) \left\{z(2\Phi(z)-1)+2\phi(z)\right\} \big\{ \left\{1-\Phi(-z)+\Phi(z) \right\}^{K-1} \bm{1}\{z<0\} + \left\{1-\Phi(z)+\Phi(-z) \right\}^{K-1} \bm{1}\{z>0\} \big\}
\end{align*}
as $\delta(x)\rightarrow 0$ for all $z$. By the dominated convergence theorem,
\begin{align*}
    \lim_{\delta(x)\rightarrow 0} \gamma(K,\delta(x)) 
    &= \int_{-\infty}^0  K\phi(z) \left\{z(2\Phi(z)-1)+2\phi(z)\right\} \left\{1-\Phi(-z)+\Phi(z) \right\}^{K-1} dz \\
    &\quad + \int_{0}^\infty  K\phi(z) \left\{z(2\Phi(z)-1)+2\phi(z)\right\} \left\{1-\Phi(z)+\Phi(-z) \right\}^{K-1} dz \\
    &= 2\int_{-\infty}^0  K\phi(z) \left\{z(2\Phi(z)-1)+2\phi(z)\right\} \left\{2\Phi(z) \right\}^{K-1} dz \\
    &= 2\left[\frac{2^K K}{K+1}\Phi(z)^{K+1}\left\{z(2\Phi(z)-1)+2\phi(z)\right\}\right]_{-\infty}^0
    -2\int_{-\infty}^0 \frac{2^K K}{K+1}\Phi(z)^{K+1}(2\Phi(z)-1) dz \\
    &= \frac{2K}{K+1} -2\int_{-\infty}^0 \frac{2^K K}{K+1}\Phi(z)^{K+1}(2\Phi(z)-1) dz \\
    &\geq \frac{2K}{K+1} \\
    &\geq 1
\end{align*}
for $K \geq 1$, which gives the first result.

Similarly, from \eqref{eq:eta},
\begin{align*}
    \eta(K,\delta(x)) &= 
    K \int_{-\infty}^{\infty} z^2 \phi(z) \big\{ \left\{1-\Phi(-2\delta(x)-z)+\Phi(z) \right\}^{K-1} \bm{1}\{z<-\delta(x)\} \\
    &\quad + \left\{1-\Phi(z)+\Phi(-2\delta(x)-z)\right\}^{K-1} \bm{1}\{z>-\delta(x)\} \big\} dz.
\end{align*}
The integrand is bounded by $K z^2 \phi(z)$, an integrable function on $\mathbb{R}$, and converges to 
\begin{align*}
    K z^2 \phi(z) \big\{ \left\{1-\Phi(-z)+\Phi(z) \right\}^{K-1} \bm{1}\{z<-0\} 
    + \left\{1-\Phi(z)+\Phi(-z)\right\}^{K-1} \bm{1}\{z>0\} \big\}
\end{align*}
as $\delta(x) \rightarrow 0$ for all $z$. By the dominated convergence theorem,
\begin{align*}
    \lim_{\delta(x)\rightarrow 0} \eta(K,\delta(x)) 
    &= \int_{-\infty}^0 K z^2 \phi(z) \left\{1-\Phi(-z)+\Phi(z) \right\}^{K-1} dz 
    + \int_0^\infty Kz^2\phi(z) \left\{1-\Phi(z)+\Phi(-z)\right\}^{K-1} dz \\
    &= 2 \int_{-\infty}^0 K z^2 \phi(z) \left\{2\Phi(z) \right\}^{K-1} dz \\
    &= 2^{K-1}\left[z^2 \Phi(z)^K\right]_{-\infty}^0 - 2^K\int_{-\infty}^0 z\Phi(z)^K dz \\
    &= -2^K\int_{-\infty}^0 z\Phi(z)^K dz \\
    &\leq - 2\int_{-\infty}^0 z\Phi(z) dz \\
    &=-\left[z^2\Phi(z)\right]_{-\infty}^0+\int_{-\infty}^0 z^2 \phi(z) dz \\
    &= \left[-z\phi(z)\right]_{-\infty}^0+\int_{-\infty}^0 \phi(z)dz \\
    &= \frac{1}{2},
\end{align*}
which concludes the proof.
\end{proof}

\subsection{Proof of Supplementary Lemmas}

Lemma \ref{lem_DPO_loss}, Lemma \ref{lemma:resample_pdf}, and Lemma \ref{lemma:DPO_loss_grad} are proved in this section.

\subsubsection{Proof of Lemma \ref{lem_DPO_loss}} \label{pf:lem_DPO_loss}

\begin{replemma}{lem_DPO_loss}
(restated) Under the DPO loss, the gradient has an alternative form
\begin{align*}
    \nabla_{\theta}\mathcal{L}(\theta) = -\mathbb{E}_{(x,y,y') \sim \mathcal{D}_u}\left[ \{\sigma(r^*(x, y) - r^*(x,y')) - \sigma(f_{\theta}(x,y) - f_{\theta}(x, y')) \}(\nabla_{\theta}f_{\theta}(x, y) - \nabla_{\theta}f_{\theta}(x,y'))\right].
\end{align*}
\end{replemma}

\begin{proof}
    First by $\sigma(x)' = \sigma(x)(1-\sigma(x))$ and $\sigma(-x) = 1 - \sigma(x)$ we observe that  
    
    \begin{equation}\label{eqn:lem_s1}
        \begin{aligned}
        \nabla_{\theta}\mathcal{L}(\theta) &= -\mathbb{E}_{(x,y_w,y_l)\sim \mathcal{D}}\left[\nabla_{\theta}\log \sigma\left(f_{\theta}(x, y_w) - f_{\theta}(x, y_l)\right)\right]\\
    &= -\mathbb{E}_{(x,y_w,y_l)\sim \mathcal{D}}\left[\sigma\left(f_{\theta}(x, y_l) - f_{\theta}(x, y_w)\right) \left(\nabla_{\theta}f_{\theta}(x, y_w) - \nabla_{\theta}f_{\theta}(x, y_l)\right)\right]\\
    &= - \int_{x}\int_{y,y'}\sigma\left(f_{\theta}(x, y') - f_{\theta}(x, y)\right) \left(\nabla_{\theta}f_{\theta}(x, y) - \nabla_{\theta}f_{\theta}(x, y')\right) p_X(x) p_{Y_w, Y_l}(y, y' | x)dy dy' dx\,. 
        \end{aligned}
    \end{equation}
Next, by a symmetrical trick, we switch the notation $y \leftrightarrow y'$, and observe

\begin{equation}\label{eqn:lem_s2}
    \begin{aligned}
    \nabla_{\theta}\mathcal{L}(\theta) &= - \int_{x}\int_{y',y}\sigma\left(f_{\theta}(x, y) - f_{\theta}(x, y')\right) \left(\nabla_{\theta}f_{\theta}(x, y') - \nabla_{\theta}f_{\theta}(x, y)\right) p_X(x) p_{Y_w, Y_l}(y', y| x)dy' dy dx\,\\
    &= - \int_{x}\int_{y,y'}\sigma\left(f_{\theta}(x, y) - f_{\theta}(x, y')\right) \left(\nabla_{\theta}f_{\theta}(x, y') - \nabla_{\theta}f_{\theta}(x, y)\right) p_X(x) p_{Y_w, Y_l}(y', y| x)dy dy' dx\,    
    \end{aligned}
\end{equation}

By combining \eqref{eqn:lem_s1} and \eqref{eqn:lem_s2} we have
\begin{equation}
    \begin{aligned}
    \nabla_{\theta}\mathcal{L}(\theta) &= -\frac{1}{2}\int_{x}\int_{y,y'}\sigma\left(f_{\theta}(x, y') - f_{\theta}(x, y)\right) p_{Y_w, Y_l}(y, y' | x) - \sigma\left(f_{\theta}(x, y) - f_{\theta}(x, y')\right)p_{Y_w, Y_l}(y', y| x)   \\
    &\quad \left(\nabla_{\theta}f_{\theta}(x, y) - \nabla_{\theta}f_{\theta}(x, y')\right) p_X(x)dy dy' dx.
    \end{aligned}
\end{equation}

Next, we analyze the term $\sigma\left(f_{\theta}(x, y') - f_{\theta}(x, y)\right) p_{Y_w, Y_l}(y, y' | x) - \sigma\left(f_{\theta}(x, y) - f_{\theta}(x, y')\right)p_{Y_w, Y_l}(y', y| x)$. From Lemma \ref{lemma_density}, we have
\begin{equation}
    \begin{aligned}
        &\hspace{5mm}\sigma\left(f_{\theta}(x, y') - f_{\theta}(x, y)\right) p_{Y_w, Y_l}(y, y' | x) - \sigma\left(f_{\theta}(x, y) - f_{\theta}(x, y')\right)p_{Y_w, Y_l}(y', y| x)\\
        &=\sigma\left(f_{\theta}(x, y') - f_{\theta}(x, y)\right)(p_{Y_1,Y_2|x}(y, y'|x) + p_{Y_1,Y_2|x}(y', y|x)) \sigma(r^*(x,y) - r^*(x,y'))\\
        &\quad - \sigma\left(f_{\theta}(x, y) - f_{\theta}(x, y')\right)(p_{Y_1,Y_2|x}(y, y'|x) + p_{Y_1,Y_2|x}(y', y|x)) \sigma(r^*(x,y') - r^*(x,y))\\
        & = \left\{\sigma\left(f_{\theta}(x, y') - f_{\theta}(x, y)\right)\sigma(r^*(x,y) - r^*(x,y')) - \sigma\left(f_{\theta}(x, y) - f_{\theta}(x, y')\right)\sigma(r^*(x,y') - r^*(x,y))\right\}\\
        &\quad \cdot (p_{Y_1,Y_2|x}(y, y'|x) + p_{Y_1,Y_2|x}(y', y|x))
    \end{aligned}
\end{equation}
Now, if we denote by 
\begin{align*}
    s &:= \sigma(r^*(x, y) - r^*(x, y'))\\
    t &:= \sigma(f_{\theta}(x, y) - f_{\theta}(x, y'))\\
    q_{Y_1,Y_2}(y,y'|x) &:= \frac{1}{2}(p_{Y_1,Y_2|x}(y, y'|x) + p_{Y_1,Y_2|x}(y', y|x)),
\end{align*}
and $\mathcal{D}_u$ the distribution of unordered pair $(y, y')$ with pmf (density) $q_{Y_1Y_2}(\cdot,\cdot|x)$, we have 
\begin{equation*}
    \begin{aligned}
    \nabla_{\theta}\mathcal{L}(\theta) &= -\int_{x}\int_{y,y'} (s(1-t) - (1-s)t)\left(\nabla_{\theta}f_{\theta}(x, y) - \nabla_{\theta}f_{\theta}(x, y')\right)q_{Y_1,Y_2}(y,y'|x)p_X(x)dy dy' dx\\
    &= -\int_{x}\int_{y,y'}(\sigma(r^*(x, y) - r^*(x, y')) - \sigma(f_{\theta}(x, y) - f_{\theta}(x, y')))\\
    &\hspace{10mm}\cdot \left(\nabla_{\theta}f_{\theta}(x, y) - \nabla_{\theta}f_{\theta}(x, y')\right)q_{Y_1,Y_2}(y,y'|x)p_X(x)dy dy' dx\\
    &= -\mathbb{E}_{(x,y,y') \sim \mathcal{D}_u}\left[\{\sigma(r^*(x, y) - r^*(x,y')) - \sigma(f_{\theta}(x,y) - f_{\theta}(x, y'))\}(\nabla_{\theta}f_{\theta}(x, y) - \nabla_{\theta}f_{\theta}(x,y'))\right],
    \end{aligned}
\end{equation*}
which concludes the proof.
\end{proof}

\subsubsection{Proof of Lemma \ref{lemma:resample_pdf}} 

\begin{replemma}{lemma:resample_pdf}
(restated) For a fixed input $x \in \mathbb{R}^d$, suppose responses $y^{(i)}$ $(i=1, \cdots, K)$ are generated by $y^{(i)} = w^\top x + \sigma \epsilon^{(i)}$ for some $w \in \mathbb{R}^d$ where $\epsilon^{(i)} \overset{iid}{\sim} \mathcal{N}(0,1)$ are Gaussian noises. Choose $m = \argmin_{i} |y^{(i)}-w^{*\top}x|$ to be the sample that is closest to the target model $y=w^{*\top}x$ and denote $\epsilon_0 := \epsilon^{(m)}$. Then the distribution of $\epsilon_0$ is 
\begin{align*}
    p_{\epsilon_0}(u) = K \phi(u) \left\{1-F_{|\delta(x)+Z|}(|\delta(x)+u|) \right\}^{K-1}
\end{align*}
where $\phi$ is the PDF of the standard normal distribution, $\delta(x) = \frac{(w-w^*)^\top x}{\sigma}$, and $F_{|\delta(x)+Z|}$ is the cdf of $|\delta(x)+Z|$ with $Z \sim \mathcal{N}(0,1)$.
\end{replemma}

\begin{proof}
By definition, $\epsilon_o = u$ if and only if $\epsilon^{(i)} = u$ for some $i$ and $|y^{(j)}-w^{*\top}x| \geq |y^{(i)}-w^{*\top}x|$ for all $j \neq i$. Note that $|y^{(j)}-w^{*\top}x| = |w^\top x + \sigma \epsilon^{(j)}-w^{*\top}x| = \sigma |\delta(x)+\epsilon^{(j)}|$.
Hence $\epsilon_0 = u$ if and only if $\epsilon^{(i)} = u$ for some $i$ and $\epsilon^{(j)} \in I$ for all $j \neq i$ where 
\begin{align*}
    I = (-\infty, -\delta(x) - |\delta(x)+u|) \cup (-\delta(x) + |\delta(x)+u|, \infty).
\end{align*} 
As $\epsilon^{(j)}$'s are iid standard normal, PDF of $\epsilon_0$ is
\begin{align}
    p_{\epsilon_0}(u) &= K \int_I \cdots \int_I \phi(z_1) \cdots \phi(z_{K-1}) \phi(u) dz_1 \cdots dz_{K-1} \nonumber \\
    &=K \phi(u) \left\{\Phi(-\delta(x) - |\delta(x)+u|) + 1-\Phi(-\delta(x) + |\delta(x)+u|) \right\}^{K-1}. \label{eq:pdf_raw}
\end{align}
Now, for $v \in \mathbb{R}$ and $Z \sim \mathcal{N}(0,1)$,
\begin{align*}
    F_{|\delta(x)+Z|}(v) &= \mathbb{P}(-v \leq \delta(x)+Z \leq v)\\
    &= \Phi(v-\delta(x)) - \Phi(-v-\delta(x))
\end{align*}
if $v\geq0$ and $F_{|\delta(x)+Z|}(v)=0$ if $v<0$. Hence
\begin{align}
    \Phi(-\delta(x) - |\delta(x)+u|) + 1-\Phi(-\delta(x) + |\delta(x)+u|) & = 1-F_{|\delta(x)+Z|}(|\delta(x)+u|). \label{eq:pdf_conversion}
\end{align}
Plugging \eqref{eq:pdf_conversion} into \eqref{eq:pdf_raw} concludes the proof.
\end{proof}

\subsubsection{Proof of Lemma \ref{lemma:DPO_loss_grad}}

\begin{replemma}{lemma:DPO_loss_grad}
(restated) Let $(x,y_+,y_-)$ be the preference data. Denote 
\begin{align*}
    \ell(\theta; x,y_+, y_-) = -\log{\tilde{\sigma}(f_\theta(x,y_+)-f_\theta(x,y_-))}
\end{align*}
where $f_\theta(x,y) = \beta \log{\frac{\pi_\theta(y|x)}{\pi_{ref}(y|x)}}$. For the Gaussian models $\pi_\theta(\cdot|x) = \mathcal{N}(w^\top x, \sigma^2)$ and $\pi_{ref}(\cdot|x) = \mathcal{N}(w_{ref}^\top x, \sigma_{ref}^2)$, the gradient and the Hessian of $l$ are 
\begin{align*}
    \nabla_w l  
    &= -\beta\frac{\sigma_{ref}}{\sigma^2} \left\{1-\tilde{\sigma}(f_\theta(x,y_+)-f_\theta(x,y_-))  \right\} (\epsilon_+ - \epsilon_-) x ,\\
    \nabla_w^2 l
    &= \beta^2 \frac{\sigma_{ref}^2}{\sigma^4 }( \tilde{\sigma} \cdot (1-\tilde{\sigma}) ) (f_\theta(x,y_+)-f_\theta(x,y_-)) \left( \epsilon_+ - \epsilon_- \right)^2 x x^\top,
\end{align*}
where $\epsilon_+ = \frac{y_+-w_{ref}^\top x}{\sigma_{ref}}$ and $\epsilon_- = \frac{y_- - w_{ref}^\top x}{\sigma_{ref}}$.
\end{replemma}

\begin{proof}
By the PDF of normal distribution, the function $f_\theta$ reduces to 
\begin{align*}
    f_\theta(x,y) = \beta \log{\sqrt{\frac{\sigma_{ref}^2}{\sigma^2}}} -\beta \frac{(y-w^\top x)^2}{2\sigma^2} + \beta \frac{(y-w_{ref}^\top x)^2}{2\sigma_{ref}^2},
\end{align*}
hence
\begin{align*}
    \frac{\partial f_\theta(x,y)}{\partial w} = \beta \frac{y-w^\top x}{\sigma^2} x.
\end{align*}
Differentiating $l$ with respect to $w$ leads to 
\begin{align}
    \frac{\partial l}{\partial w} 
    &= -\left\{1-\tilde{\sigma}(f_\theta(x,y_+)-f_\theta(x,y_-))  \right\} \left(\frac{\partial f_\theta(x,y_+)}{\partial w} - \frac{\partial f_\theta(x,y_-)}{\partial w}\right) \nonumber \\
    &= -\left\{1-\tilde{\sigma}(f_\theta(x,y_+)-f_\theta(x,y_-))  \right\} \beta \left( \frac{y_+ - w^\top x}{\sigma^2} - \frac{y_- - w^\top x}{\sigma^2}\right) x  \label{eq:grad2} \\
    &= -\beta\frac{\sigma_{ref}}{\sigma^2} \left\{1-\tilde{\sigma}(f_\theta(x,y_+)-f_\theta(x,y_-))  \right\} (\epsilon_+ - \epsilon_-) x. \nonumber
\end{align}
Also, from \eqref{eq:grad2},
\begin{align*}
    \frac{\partial^2 l}{\partial w^2} 
    &= ( \tilde{\sigma} \cdot (1-\tilde{\sigma}) ) (f_\theta(x,y_+)-f_\theta(x,y_-)) \beta^2\left( \frac{y_+ - w^\top x}{\sigma^2} - \frac{y_- - w^\top x}{\sigma^2}\right)^2 x x^\top  \\
    &\quad -\left\{1-\tilde{\sigma}(f_\theta(x,y_+)-f_\theta(x,y_-))  \right\} \beta \left( -\frac{ 1}{\sigma^2}x + \frac{1}{\sigma^2 }x\right) x^\top  \\
    &= \beta^2 \frac{\sigma_{ref}^2}{\sigma^4 }( \tilde{\sigma} \cdot (1-\tilde{\sigma}) ) (f_\theta(x,y_+)-f_\theta(x,y_-)) \left( \epsilon_+ - \epsilon_- \right)^2 x x^\top. 
\end{align*}
\end{proof}

\end{document}

%% file: math_commands.tex
\usepackage{amsmath,amsfonts,bm}

\def\eqref#1{equation~\ref{#1}}

\def\1{\bm{1}}

\DeclareMathAlphabet{\mathsfit}{\encodingdefault}{\sfdefault}{m}{sl}
\SetMathAlphabet{\mathsfit}{bold}{\encodingdefault}{\sfdefault}{bx}{n}

\DeclareMathOperator*{\argmax}{arg\,max}
\DeclareMathOperator*{\argmin}{arg\,min}